\documentclass{neus2025}

\title[Short Title]{Full Title of Article}
\usepackage{times}

\author{%
 \Name{William English} \Email{will.english@ufl.edu}\\
  \Name{Hao Zheng} \Email{hao.zheng@ucf.edu}\\
 \Name{Rickard Ewetz} \Email{rewetz@ufl.edu}\\
}

\usepackage{wrapfig}
\usepackage{amsmath,amssymb,amsfonts}
\usepackage{algorithmic}
\usepackage{algorithm}
\usepackage{graphicx}
\usepackage{textcomp}
\usepackage{booktabs}
\usepackage{hyperref}
\usepackage{multirow}
\usepackage{subcaption}
\usepackage{soul}

\begin{document}
\title[Neuro-Symbolic Safety Guidance for VLAs via Constrained Flow Matching]{Neuro-Symbolic Safety Guidance for Vision-Language-Action Models \\ via Constrained Flow Matching}



\maketitle


\begin{abstract}
Vision-Language-Action (VLA) models have demonstrated promising generalization capabilities across robotic manipulation tasks, yet their real-world deployment remains limited by the lack of effective safety measures. Specifically, existing safety measures only prevent collisions caused by the robot’s next action. In this paper, we propose a neuro-symbolic safety guidance mechanism for flow matching based VLAs that enables predictive collision avoidance.
Flow matching based VLAs determine the next actions by predicting a trajectory (a sequence of actions) through an iterative neural flow matching process. Our method formulates safety enforcement as a minimum-norm constrained optimization problem that corrects safety violations during the denoising process of noisy intermediate trajectory predictions. By analyzing predicted trajectories and applying corrections during iterative denoising, our approach anticipates collisions before they become unavoidable. This interleaving of symbolic constraint satisfaction with neural trajectory generation enables predictive collision avoidance rather than reactive intervention.
On the SafeLIBERO benchmark, our method achieves 82.8\% collision avoidance and 81.6\% task success, a  6.3\% and 19.8\%  improvement respectively over single-step methods, with the largest gains on long-horizon tasks where compounding distribution shift is most pronounced.
Video demonstrations of our approach are included on our project page at \href{https://willenglish.tech/SafetyGuidedFlowMatching/}{https://willenglish.tech/SafetyGuidedFlowMatching/}.
\end{abstract}


\section{Introduction}
\label{sec:introduction}

Vision-Language-Action (VLA) models represent a significant advance toward general-purpose robotic manipulation, unifying visual perception, language understanding, and action generation within a single end-to-end framework (\cite{rt2_vla, openvla}). Recent models such as $\pi_{0.5}$ (\cite{pi0-paper, pi05-paper}) have demonstrated impressive generalization across diverse manipulation tasks through flow matching-based action generation. However, deploying these models in real-world environments requires addressing a fundamental challenge: ensuring physical safety, particularly collision avoidance, during task execution. Collisions during manipulation can both disrupt task execution and damage the robotics hardware.

Existing approaches to VLA safety fall into two categories. Training-time methods enforce safety through reinforcement learning~\citep{safevlazhang2025}, requiring costly retraining and treating safety as a soft objective rather than a hard constraint. Inference-time approaches such as AEGIS and SafeDec~\citep{vlsa, safedec} instead couple the neural policy with an explicit, declarative control barrier function (CBF) defining a safe set, enforced by a solver rather than learned. However, these methods place the symbolic solver \emph{outside} generation, projecting a finished action onto the constraint surface post-hoc, whereas decode-time enforcement in neuro-symbolic constrained generation interleaves the solver \emph{within} the generative loop~\citep{lu-etal-2021-neurologic}. Because post-hoc correction operates on individual actions rather than trajectories, it cannot anticipate violations until imminent, forcing large last-moment interventions that disrupt task
execution.

In this paper, we propose a neuro-symbolic safety guidance mechanism for flow matching VLAs that enables predictive collision avoidance. Flow matching based VLAs produce action trajectories through iterative denoising over multiple Euler integration steps. At each step, we treat the current action trajectory as a predicted trajectory, evaluate discrete-time CBF constraints against obstacle geometries to predict collisions, and solve a minimum-norm constrained optimization to correct violations. The corrected trajectory serves as input to the next denoising step, where the velocity field further reduces noise in the trajectory prediction. This feedback loop between constraint satisfaction and neural refinement enables predictive, trajectory-level safety. The system identifies collisions beyond the immediate future and distributes corrections to earlier actions, initiating avoidance while obstacles are still distant. Our proposed trajectory-based approach is contrasted with no and single action collision avoidance in  Figure~\ref{fig:trajectory_distributions}. Our contributions are summarized, as follows:

\begin{figure}
 \vspace{-6pt}
    \centering
    \subfigure[Unguided]{\includegraphics[width=0.31\linewidth]{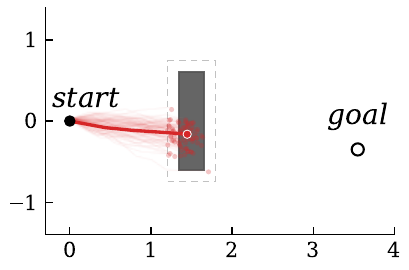}}
    \label{subfig:trajectory_distributionsa}
    \subfigure[Single-Step]{\includegraphics[width=0.31\linewidth]{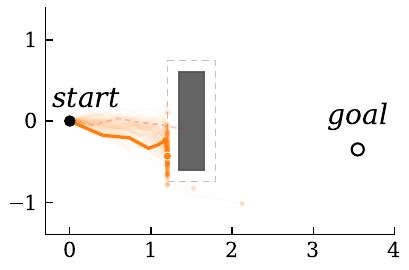}}
    \label{subfig:trajectory_distributionsb}
    \subfigure[Full-trajectory (ours)]{\includegraphics[width=0.31\linewidth]{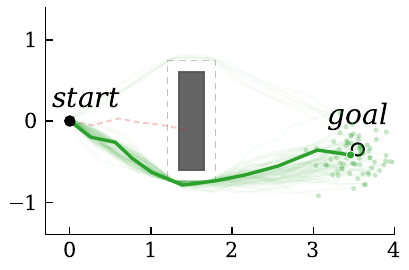}}
    \label{subfig:trajectory_distributionsc}
        \caption{Path distributions ($N=100$ samples) for a point robot navigating past a rectangular obstacle. (a) A VLA without safety guidance produces trajectories that collide with the obstacle. (b) Single-action CBF filtering avoids collisions by intervening when a collision would occur in the next action, often resulting in excessive detours or deadlock failures~\citep{predictiveLi2025} (c) The proposed safety guidance operates over entire trajectories, enabling anticipatory avoidance.}
    \label{fig:trajectory_distributions}
    \vspace{-6pt}
\end{figure}

\begin{enumerate}
    \item \textbf{Trajectory-level safety insight:} We observe that the flow matching action generation process in VLAs can be interpreted as iterative trajectory prediction, enabling predictive collision avoidance by reasoning over future motion rather than single actions.

    \item \textbf{Neuro-symbolic safety guidance during generation:} We propose a neuro-symbolic safety guidance mechanism that interleaves symbolic barrier-constraint evaluation and minimum-norm trajectory correction with neural flow matching denoising. This iterative neuro-symbolic process converges to action predictions that satisfy safety constraints while remaining close to the model’s intended behavior.

    \item \textbf{Evaluation:} On the SafeLIBERO benchmark, our method achieves 82.8\% collision avoidance and 81.6\% task success, improving over single-action CBF filtering (77.9\% and 68.1\%, respectively), with the largest gains observed on long-horizon tasks.

\end{enumerate}

The remainder of this paper is organized as follows. In Section \ref{sec:preliminaries}, we discuss preliminaries about VLAs and safe autonomy. In Section \ref{sec:related_safety_in_robot}, we discuss related work including the SafeLIBERO benchmark and VLA-specific safety methods. The details of our methodology are explained in Section  \ref{sec:method}. Our experimental evaluation and conclusions are presented in Section \ref{sec:experiments} and \ref{sec:conclusion}, respectively.
\section{Preliminaries}
\label{sec:preliminaries}

\subsection{Vision-Language-Action Models}

VLA models integrate visual perception and language understanding for robotic action generation within end-to-end architectures. RT-2 \citep{rt2_vla} pioneered this paradigm by co-fine-tuning vision-language models on robotic trajectories, expressing actions as text tokens. OpenVLA \citep{openvla} extended this as an open-source 7B-parameter model trained on 970k real-world demonstrations, enabling efficient fine-tuning via low-rank adaptation. The $\pi_0$ and $\pi_{0.5}$ models \citep{pi0-paper, pi05-paper} introduced flow matching for continuous action generation, learning velocity fields that transform noise into action trajectories through iterative denoising. Their approach has been widely adopted by the community, inspiring a number of adjacent flow matching approaches \citep{zhang2024flowpolicy, gao2025vita, yan2025maniflow, jeon2026shallowpiknowledgedistillationflowbased}

This iterative structure connects to guided generation methods that inject external objectives into diffusion or flow matching sampling. Classifier-free guidance \citep{ho2022classifierfreediffusionguidance} steers sampling using learned conditional signals, while Diffuser \citep{janner2022planningdiffusionflexiblebehavior} applied diffusion-based trajectory optimization, using gradient-based guidance to satisfy task objectives during sampling.

\subsection{Flow Matching for Action Generation}
\label{sec:flow_matching}
Flow matching VLAs generate action chunks via a learned velocity field that transports samples from noise to the action distribution (\cite{flow-matching}) . Given an observation $o_t$ comprising RGB images, a language command, and proprioceptive state, the model generates an action chunk $A_t = [a_t, a_{t+1}, \ldots, a_{t+H-1}]$ of $H$ future actions (typically $H=10$). At inference, actions are produced by integrating a learned velocity field $v_\theta$ from flow time $\tau = 1$ (noise) to $\tau = 0$ (actions). Starting from random noise $A_t^1 \sim \mathcal{N}(0, I)$, forward Euler integration yields:
\begin{equation}
\label{eq:euler_step}
    A_t^{\tau + \Delta \tau} = A_t^\tau + \Delta \tau\cdot v_\theta(A_t^\tau, o_t)
\end{equation}
where $\Delta \tau = -1/N$ is the step size for $N$ integration steps (typically $N = 10$). The final action chunk $A_t^0$ is then executed on the robot.
Each action $a_i \in \mathbb{R}^d$ specifies end-effector velocities and a gripper command. 
Executing the chunk through the robot dynamics produces a trajectory of states $\{s_{0}, s_{1}, \ldots, s_{H}\}$, where $s_{0}$ is the current state and each transition follows $s_{i+1} = f(s_{i}, a_i)$. This trajectory is the object over which we enforce safety constraints. 

\subsection{Control Barrier Functions}
\label{sec:cbf_prelim}
Control Barrier Functions (CBFs) \citep{CBF-Theory, CBF-Safety} provide a symbolic framework for enforcing safety in autonomous systems by guaranteeing forward invariance of a designated safe set. Consider a discrete-time system $s_{t+1} = f(s_t, a_t)$ where $s_t \in \mathcal{S}$ is the state, $a_t \in \mathcal{A}$ is the control input, and $f: \mathcal{S} \times \mathcal{A} \rightarrow \mathcal{S}$ is a continuous map describing the system dynamics. Given a continuous function $B: \mathbb{R}^n \rightarrow \mathbb{R}$, define the safe set $\mathcal{C} = \{x : B(x) \geq 0\}$, its boundary $\partial\mathcal{C} = \{x : B(x) = 0\}$, and its interior $\texttt{Int}(\mathcal{C}) = \{x : B(x) > 0\}$.

$B$ is a discrete-time exponential control barrier function \citep{Agrawal2017DiscreteCB} for the system if, for all $s_t \in \mathcal{C}$, there exists a control input $a_t$ such that:
\begin{equation}
    \Delta B(s_t, a_t) + \gamma\, B(s_t) \geq 0, \quad \gamma \leq 1
    \label{eq:dcbf_condition}
\end{equation}
where $\Delta B(s_t, a_t) := B(s_{t+1}) - B(s_t)$. Rearranging yields the equivalent condition $B(s_{t+1}) \geq (1-\gamma)\, B(s_t)$: the barrier value at the next state must remain above a fraction $(1-\gamma)$ of the current value. When $\gamma = 1$, this reduces to $B(s_{t+1}) \geq 0$, requiring safety at every step. Smaller $\gamma$ permits gradual approach toward the boundary, producing smoother corrections. Given $B(s_0) \geq 0$, this condition guarantees $B(s_t) \geq (1-\gamma)^t\, B(s_0) \geq 0$ for all $t$, ensuring the system remains within $\mathcal{C}$.

In most settings, only a single nominal control input $a_{\text{t}}$ is available at time $t$. This action is filtered by solving for the minimum-norm modification satisfying \eqref{eq:dcbf_condition}. This formulation evaluates the barrier at the current state and constrains only the immediate next state. Our method extends this notion to settings in which a trajectory of actions $\mathcal{A}_{t,t+H}$ (where H is the action horizon of a predictive controller) is known.

\subsection{Problem Formulation}
\label{subsec:problem_formulation}
We consider a robotic manipulation setting where a VLA policy generates an action chunk $A=[a_1, \ldots, a_H]$ of horizon $H$ in response to a natural language instruction and stream of visual observations. Execution of this action sequence from initial state $s_0$ produces a trajectory $\{s_0, s_1, \ldots, s_H\}$ according to the dynamics $s_{i+1} = f(s_i, a_i)$. The environment contains a set of obstacles $\mathcal{O} = \{O_1, \ldots, O_K\}$ with known geometry and positions. An example of an environment initialization is shown below. We model the robot end-effector as an ellipsoid with semi-axes $r = (r_x, r_y, r_z)$. 

\begin{wrapfigure}[]{r}{0.35\linewidth}
  \begin{center}
  \vspace{-25.0pt}
  \includegraphics[width=\linewidth]{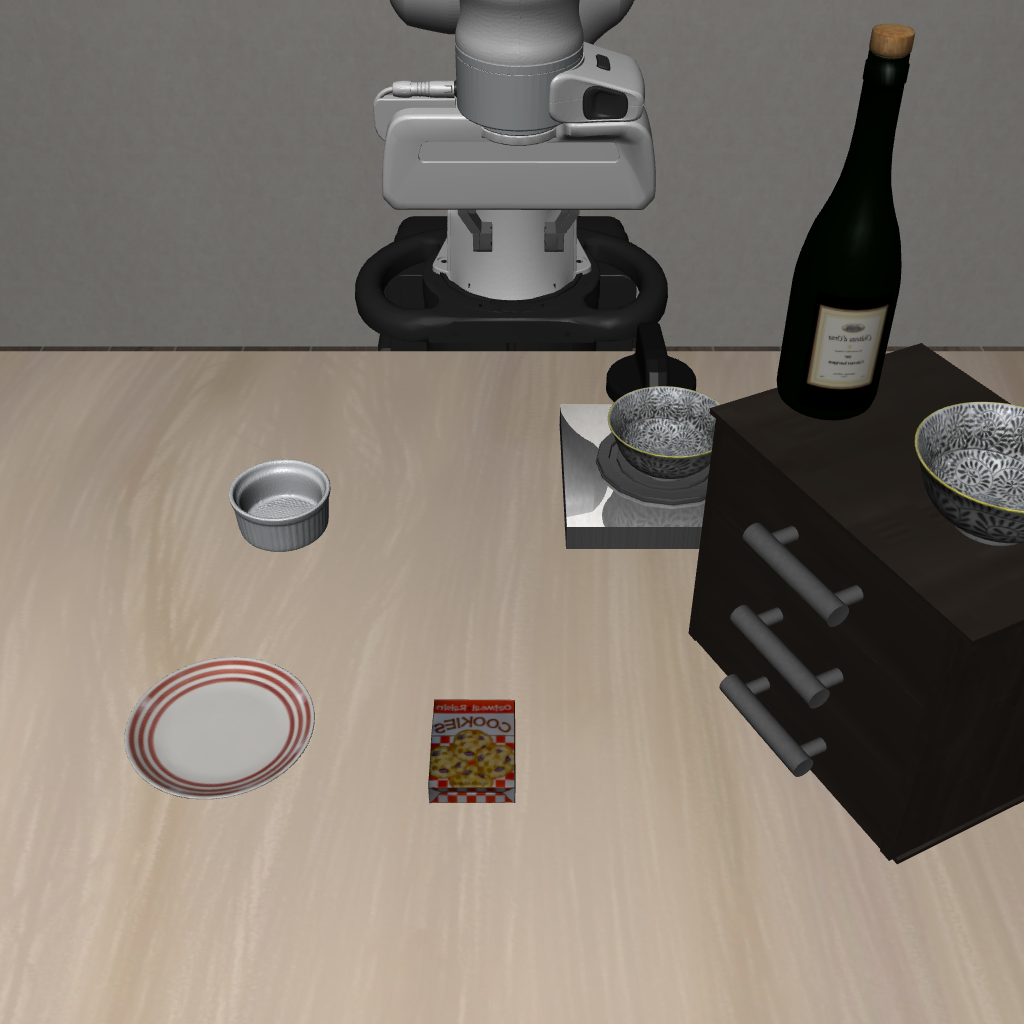}
  {\centering\scriptsize In this SafeLIBERO task, the wine bottle obstructs the arm from picking up the bowl.}
  
    \vspace{-45.0pt}

  \end{center}
\end{wrapfigure}

For each state $s_i$ and obstacle $O_j$, we define a barrier function $B_{ij}(s_i)$ that is positive when the end-effector maintains at least $d_{safe}$ clearance from $O_j$ and negative otherwise. A trajectory is safe if:
\begin{equation}
    B_{ij}(s_i) \geq 0 \quad \forall\, i \in \{0, \ldots, H\},\; 
    j \in \{1, \ldots, K\}
    \label{eq:safety_condition}
\end{equation}
The objective is to successfully complete the natural language task while avoiding collisions. Both task completion and collision detection are determined automatically by the terminal position of objects in the simulation.

\section{Related Work on Safety in Autonomous Control}
\label{sec:related_safety_in_robot}
Classical inference-time approaches to trajectory safety, including A* and RRT* \citep{mincostpaths, sampling-motion-planning}, provide path-level safety but are difficult to integrate with end-to-end learned policies. Training-time approaches such as SafeVLA \citep{safevlazhang2025} instead incorporate safety into vision-language-action models via constrained reinforcement learning, but require expensive retraining and treat safety as a soft objective subject to reward trade-offs. AEGIS \citep{vlsa} avoids retraining by applying CBF-based quadratic programming to VLA outputs, minimally adjusting each action to satisfy safety constraints in real time. However, this post-hoc approach operates independently of the generative process: the model produces an action, and a separate symbolic layer projects it onto the constraint surface. Because these corrections operate without knowledge of future actions, they inherit the myopia of standard CBF methods \citep{cohen2020myopic, garg2024cbf}, which enforce safety only at the current timestep and cannot prevent a system from entering states where future safety is unreachable, leading to infeasibility or deadlock (Figure~\ref{fig:trajectory_distributions}).

Recent methods address this problem by embedding CBF-style constraints directly into flow matching, enforcing safety across the full generated trajectory rather than a single action \citep{dai2025safeflowsaferobotmotion, yang2026safeflowmatcher}. These are the closest to ours in mechanism, but they are task-specific motion planners trained per task to generate state-space trajectories conditioned on a start/goal pair. In contrast, we enforce trajectory-level safety on a general pretrained vision-language-action policy ($\pi_{0.5}$) that maps perception and language to action chunks, correcting in action space during generation so that subsequent denoising remains consistent with the learned velocity field.

\section{Method}
\label{sec:method}

\begin{figure}[t]
    \centering
    \includegraphics[width=\linewidth]{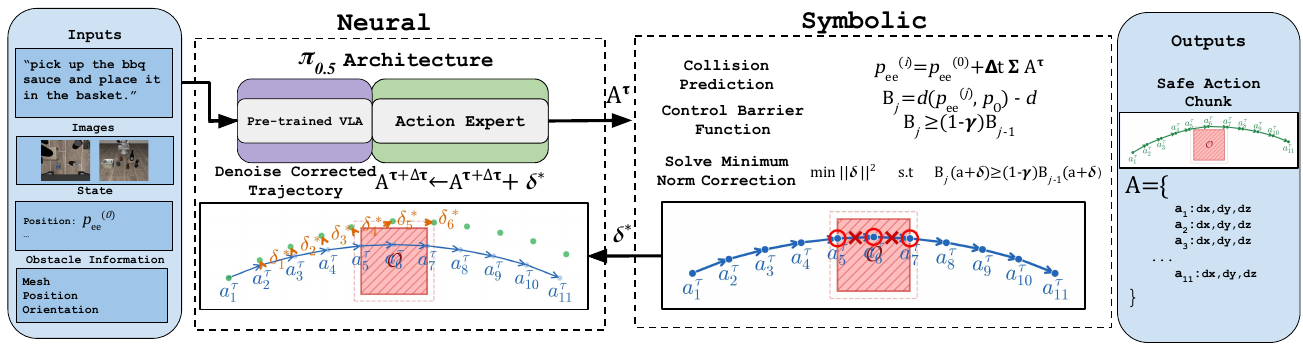}
    \caption{Overview of safety-guided flow matching. During iterative denoising, we predict the trajectory implied by the current action chunk estimate, identify future collisions with obstacles, and apply localized corrections to actions preceding each violation. 
    }
    \label{fig:method_overview}
\end{figure}

Given a VLA policy that generates actions chunks through flow matching, our objective is to ensure the resulting trajectories avoid collisions without sacrificing task completion capabilities. We propose a predictive safety guidance mechanism that integrates into the flow matching action generation process of $\pi_{0.5}$ through two key components: (1) collision prediction (Section \ref{subsec:collision_prediction}), which uses action chunks to estimate future safety violations, and (2) safety guidance (Section \ref{subsec:safety_guidance}), which computes minimum-norm corrections to action chunks that restore safety across the full predicted trajectory. An overview of our approach is illustrated in 
Figure~\ref{fig:method_overview}. At each denoising step, we take as input the current action chunk $A^\tau$ and physical robot state $p_{ee}$ to check for collisions between the end-effector and known obstacles $\mathcal{O}$. The collision prediction component computes future end-effector positions $\{p_{\text{ee}}^{(j)}\}_{j=0}^{H}$ using forward kinematics given the action chunk. When violations of the safety condition are detected, the safety guidance component solves a constrained optimization problem to find the smallest correction $\delta^*$ to the action chunk that satisfies the trajectory-level constraints. These corrections are then applied to $A^\tau$ before the next denoising step. The result of this process is the generation of safe action trajectories that are minimally offset from the unguided trajectory.

\subsection{Collision Modeling and Trajectory Prediction}
\label{subsec:collision_prediction}
We model the robot end-effector as an ellipsoid with semi-axes $r = (r_x, r_y, r_z)$ that approximate the gripper geometry, and obstacles as spheres with position $p_{\text{obs}}$ and effective radius $r_{\text{obs}}$.

\paragraph{Ellipsoid-to-obstacle signed distance.} Given the end-effector centered at position $p$, we compute an approximate signed distance to an obstacle in two steps. First, we map the obstacle position into a normalized coordinate frame where the ellipsoid becomes a unit sphere:
\begin{equation}
    \tilde{q} = (p_{\text{obs}} - p) \oslash r
    \label{eq:normalized_coords}
\end{equation}
where $\oslash$ denotes element-wise division by the semi-axes. The norm $\|\tilde{q}\|$ acts as a generalized distance: $\|\tilde{q}\| > 1$ means the obstacle center lies outside the ellipsoid, and $\|\tilde{q}\| < 1$ means it lies inside. We use this to find the point on the ellipsoid surface along the direction toward the obstacle:
\begin{equation}
    p_{\text{surf}} = p + \frac{\tilde{q}}{\|\tilde{q}\|} \odot r
    \label{eq:surface_point}
\end{equation}
The signed distance between the two bodies is then the Euclidean gap between this surface point and the obstacle center, minus the obstacle radius, with sign determined by whether the obstacle center is inside or outside the ellipsoid:
\begin{equation}
    d(p, p_{\text{obs}}) = \bigl(\|p_{\text{obs}} - p_{\text{surf}}\| \cdot \operatorname{sign}(\|\tilde{q}\| - 1)\bigr) - r_{\text{obs}}
    \label{eq:signed_distance}
\end{equation}
Positive values indicate separation; negative values indicate penetration. While the true closest point on an ellipsoid requires solving a quartic, this approximation is cheap to evaluate and differentiable, which is important since we evaluate it $H+1$ times per action chunk at every denoising step.

\paragraph{Trajectory prediction.} A key advantage of action chunk prediction is that we obtain not just the immediate action but a sequence of $H$ future actions ($H = 10$ for the $\pi_{0.5}$ LIBERO checkpoint). Given the current end-effector position $p_{\text{ee}}^{(0)}$ and an action chunk where each action specifies end-effector velocities, we predict future positions by cumulative integration:
\begin{equation}
    p_{\text{ee}}^{(i)} = p_{\text{ee}}^{(0)} + \Delta t \sum_{k=1}^{i} a_k[1{:}3]
    \label{eq:trajectory_prediction}
\end{equation}
where $a_k[1{:}3]$ extracts the translational velocity components from action $k$ and $\Delta t$ is the control timestep. If no predicted position violates the safety margin, the action chunk proceeds unchanged.

\begin{figure}
    \centering
    \includegraphics[width=\linewidth]{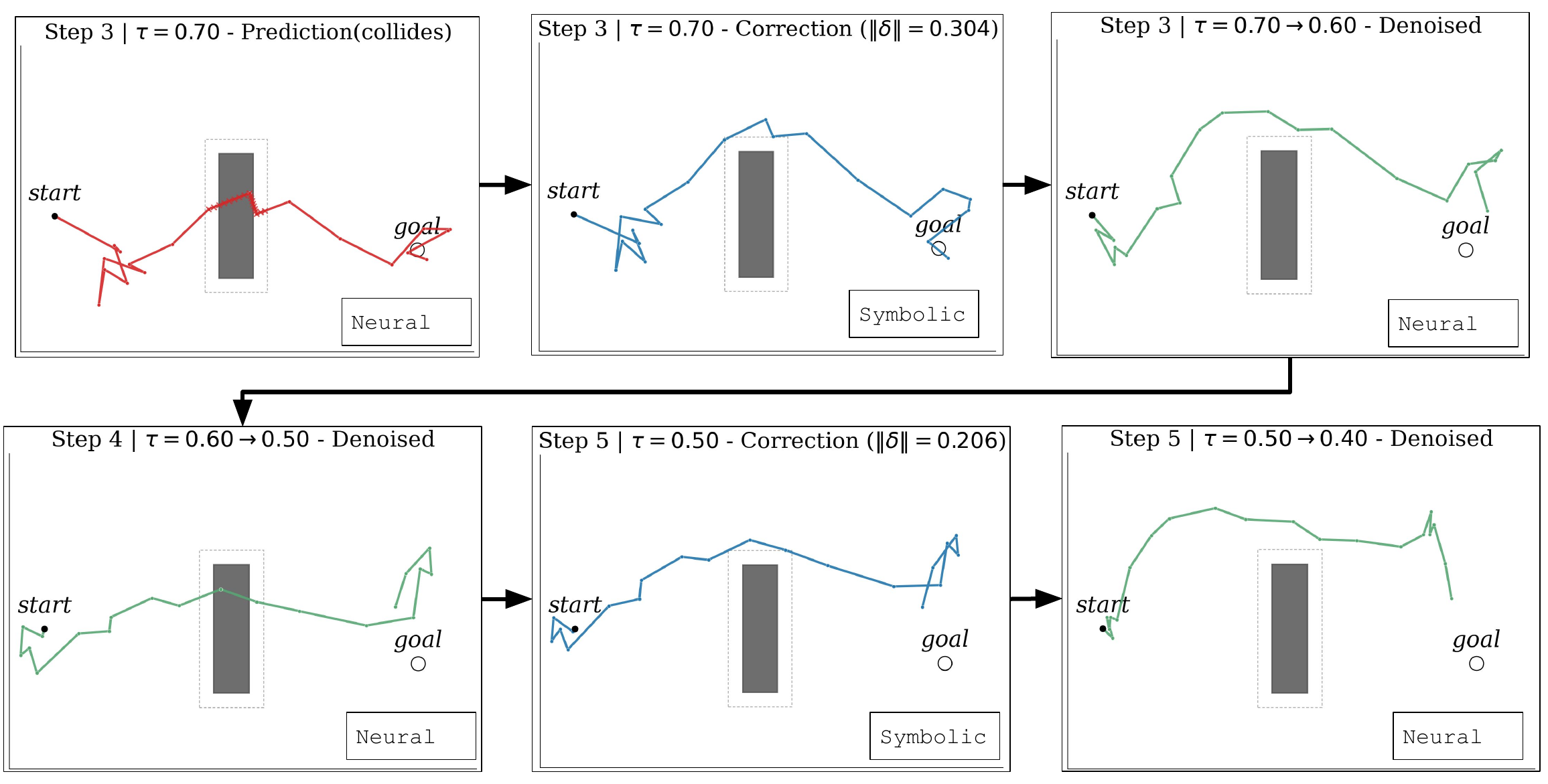}
    \caption{Steps 3-5 out of 10 of neural trajectory prediction and symbolic correction. Action trajectories are corrected and used as input at each denoising step. The complete algorithm is given in Appendix \ref{app:alg}.}
    \label{fig:placeholder}
\end{figure}

\subsection{Predictive Safety Guidance via Control Barrier Functions}
\label{subsec:safety_guidance}

The standard CBF formulation (Section~\ref{sec:cbf_prelim}) filters a single nominal action by enforcing the barrier condition \eqref{eq:dcbf_condition} at the current state. Our method extends this in two ways. First, we enforce the barrier condition across the \emph{entire predicted trajectory} implied by the action chunk, not just at the current timestep. Second, we apply the resulting corrections \emph{during} the flow matching denoising process rather than after generation, allowing subsequent neural refinement steps to adapt to the corrected trajectory.

\paragraph{Barrier function instantiation.} We instantiate the barrier function $B$ from Section \ref{sec:cbf_prelim} using the signed distance 
\eqref{eq:signed_distance}. For each predicted trajectory point $j \in \{0, \ldots, H\}$ 
and obstacle $O_k \in \mathcal{O}$, we define:
\begin{equation}
    B_j = d(p_{\text{ee}}^{(j)},\; p_{O_k}) - d_{\text{safe}}
    \label{eq:barrier}
\end{equation}

where $d_{\text{safe}}$ is the safety margin. The system is safe at point $B_j$ when $B_j \geq 0$, i.e., the ellipsoidal end-effector maintains at least $d_{\text{safe}}$ clearance from the obstacle surface.

\paragraph{Trajectory-level CBF constraint.} We enforce the discrete-time exponential CBF condition \eqref{eq:dcbf_condition} between every pair of consecutive trajectory points:
\begin{equation}
    B_j \geq (1 - \gamma)\, B_{j-1}, \quad j = 1, \ldots, H
    \label{eq:cbf_constraint}
\end{equation}
where $\gamma \in (0, 1]$ governs how aggressively safety is enforced. When $\gamma = 1$, the constraint reduces to $B_j \geq 0$: every trajectory point must be safe regardless of its predecessor. Smaller values of $\gamma$ permit gradual approach toward the obstacle boundary, producing smoother corrections.

The critical difference from standard CBF filtering is the scope of enforcement. Where \eqref{eq:dcbf_condition} constrains a single transition $(s_t, s_{t+1})$, we apply it across the full sequence $j = 1, \ldots, H$. If the current action chunk estimate would produce a violation at step $j$, we identify and correct it now, even though the robot has not yet reached $p_{\text{ee}}^{(j)}$. This predictive evaluation is possible precisely because flow matching produces action chunks rather than single actions, giving us a trajectory to inspect at each denoising step.

\paragraph{Minimum-norm correction via constrained optimization.} When any constraint in~\eqref{eq:cbf_constraint} is violated, we find the smallest modification to the action chunk that restores safety. Let $\delta \in \mathbb{R}^{H \times 3}$ denote a correction to the translational components of the action chunk. We solve:
\begin{equation}
    \min_{\delta} \;\|\delta\|^2 \quad \text{s.t.} \quad B_j(\mathbf{a} + \delta) \geq (1-\gamma)\, B_{j-1}(\mathbf{a} + \delta), \;\; j = 1, \ldots, H
    \label{eq:cbf_qp}
\end{equation}
The objective minimizes total deviation from the VLA's predicted action chunk. Because the barrier function \eqref{eq:barrier} depends nonlinearly on positions (via the signed distance \eqref{eq:signed_distance}), we solve \eqref{eq:cbf_qp} using sequential least-squares programming (SLSQP). For $H=10$, this is a 30-variable problem with 10 inequality constraints, solving in under a millisecond.

\paragraph{Injection during generation.} We modify the flow matching Euler step from \ref{eq:euler_step} with the corrected action chunk:
\begin{equation}
    A^{\tau+\Delta\tau} \gets A^{\tau+\Delta\tau} + \delta^* \quad
\end{equation}
where $\delta^*$ is the solution to \eqref{eq:cbf_qp}. Because the flow matching model receives the corrected trajectory at the next denoising step, it adapts its velocity field prediction to accommodate the safety correction. Subsequent refinement absorbs the perturbation, producing an action chunk that satisfies the barrier constraints while remaining consistent with the learned velocity field. Post-hoc methods lack this feedback: they filter each action after generation is complete, with no opportunity for the model to adapt.

In Appendix \ref{app:alg} we include Algorithm~\ref{alg:guided_flow} which summarizes the complete procedure, and Table~\ref{tab:hyperparams} which lists the parameters used during evaluation.


\begin{wrapfigure}[]{r}{0.35\linewidth}
\vspace{-75.0pt}
  \begin{center}
  \includegraphics[width=\linewidth]{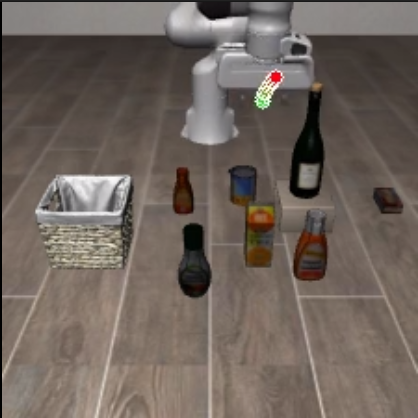}
    \caption{Example of predictive collision avoidance on a SafeLIBERO-Object task. The red point represents the predicted position of the end effector 10 time-steps in the future.}
  \end{center}
      \label{fig:collision_avoidance}
      \vspace{-50.0pt}
\end{wrapfigure}

\section{Experimental Evaluation}
\label{sec:experiments}
We evaluate on SafeLIBERO \citep{vlsa}, a safety-critical benchmark derived from the LIBERO dataset \citep{libero}. SafeLIBERO introduces obstacles into manipulation scenarios at two difficulty levels: Level I places obstacles near target objects, while Level II positions obstacles along the robot's movement path. The benchmark spans four task suites: Spatial, Object, Goal, and Long. Examples from each safety level and task suite can be found in Appendix \ref{app:safelibero}. We also provide a set of videos comparing our approach to the unguided policy on our \href{https://willenglish.tech/SafetyGuidedFlowMatching/}{project page}.

\subsection{Experimental Setup}
 We evaluate our approach against \textbf{$\pi_{0.5}$}-$\text{LIBERO}$ (\cite{pi05-paper}), the base VLA model without safety mechanisms, and AEGIS \citep{vlsa}, which is described in Section \ref{sec:related_safety_in_robot}. For each task suite, we report the following metrics: {Collision Avoidance Rate (CAR)}: percentage of episodes with no collisions; {Task Success Rate (TSR)}: percentage of episodes completing the task; {Execution Time-steps (ETS)}: mean number of time steps until episode completion.

We use the $\pi_{0.5}$-$\text{LIBERO}$ checkpoint as the base VLA model. Obstacle positions are obtained from the simulator, isolating the control contribution from perception challenges. The end-effector is modeled as an ellipsoid with dimensions matching the gripper geometry. 

\subsection{Main Results}

We present results averaged across all SafeLIBERO tasks and safety levels in Table~\ref{tab:main_results}. Our method achieves 82.81\% CAR and 81.62\% TSR overall, improving over AEGIS on both metrics (77.85\% CAR, 68.13\% TSR). The base $\pi_{0.5}$ model without safety mechanisms achieves only 18.69\% CAR, confirming that explicit safety enforcement is necessary in cluttered environments.

The most notable difference between our method and AEGIS is in task success rate. While AEGIS improves CAR substantially over the baseline, its TSR gains are more modest (68.13\% vs.\ 50.88\%), and on Long-horizon tasks its TSR of 43.75\% remains low, suggesting that post-hoc corrections disrupt the policy's ability to complete multi-step tasks. Our method achieves 76.75\% TSR on Long tasks, nearly doubling AEGIS's performance on this suite while also improving CAR (82.50\% vs.\ 79.63\%). This is consistent with the safety-induced distribution shift hypothesis discussed in Section~\ref{sec:method}: by integrating corrections into the generative process rather than applying them post-hoc, subsequent denoising steps can adapt to safety adjustments, preserving coherence with the learned action distribution.

Across suites, our method achieves the highest CAR on Goal (88.25\% vs.\ 81.50\%), Object (84.00\% vs.\ 74.75\%), and Long (82.50\% vs.\ 79.63\%), while performing comparably on Spatial (76.50\% vs.\ 75.50\%). Task success rates follow a similar pattern, with our method leading on all four suites. The primary trade-off is execution time. Our method averages 299.97 ETS compared to 262.30 for AEGIS, with the gap most pronounced on Object tasks (305.62 vs.\ 201.26). This likely reflects that trajectory-level guidance produces more conservative paths around obstacles, trading directness for safety. On Spatial tasks, however, our ETS (186.35) is comparable to AEGIS (188.20), suggesting that when guidance corrections are small, the method does not impose meaningful overhead.

\begin{table}[ht]
\small
\centering
\caption{Quantitative results on the SafeLIBERO benchmark.}
\label{tab:main_results}
\begin{tabular}{llcccc|c}
\toprule
\multirow{2}{*}{\textbf{Method}} & \multirow{2}{*}{\textbf{Metric}} & \multicolumn{4}{c|}{\textbf{SafeLIBERO Suite}} & \multirow{2}{*}{\textbf{Average}} \\
\cmidrule{3-6}
& & Spatial & Goal & Object & Long & \\
\midrule
\multirow{3}{*}{$\pi_{0.5}$} 
& CAR ($\uparrow$) & 15.25\% & 23.75\% & 23.00\% & 12.75\% & 18.69\% \\
& TSR ($\uparrow$) & 59.75\% & 54.25\% & 53.75\% & 35.75\% & 50.88\% \\
& ETS ($\downarrow$) & 201.65 & 210.31 & 223.01 & \textbf{477.99} & 278.24 \\
\midrule
\multirow{3}{*}{AEGIS} 
& CAR ($\uparrow$) & 75.50\% & 81.50\% & 74.75\% & 79.63\% & 77.85\% \\
& TSR ($\uparrow$) & 73.25\%& 75.25\% & 80.25\% & 43.75\% & 68.13\% \\
& ETS ($\downarrow$) & 188.20 & 179.60 & 201.26 & 480.12 &\textbf{ 262.30} \\
\midrule
\multirow{3}{*}{\textbf{Ours}} 
& CAR ($\uparrow$) & \textbf{76.50\%} & \textbf{88.25\%} & \textbf{84.00\%} & \textbf{82.50\%} & \textbf{82.81\%} \\
& TSR ($\uparrow$) & \textbf{76.75\%} & \textbf{87.25}\% & \textbf{85.75\%} & \textbf{76.75\%} & \textbf{81.62\%} \\
& ETS ($\downarrow$) & \textbf{186.35} & 207.55 & 305.62 & 500.35 & 299.97 \\
\bottomrule
\end{tabular}
\end{table}

\begin{table}[ht]
\centering
\caption{Quantitative results by task suite on the SafeLIBERO benchmark.}
\label{tab:suite_results}
\resizebox{\linewidth}{!}{
\begin{tabular}{ll cc cc cc cc | cc}
\toprule
& & \multicolumn{2}{c}{\textbf{Spatial}} & \multicolumn{2}{c}{\textbf{Goal}} & \multicolumn{2}{c}{\textbf{Object}} & \multicolumn{2}{c|}{\textbf{Long}} & \multicolumn{2}{c}{\textbf{Overall}} \\
\cmidrule(lr){3-4} \cmidrule(lr){5-6} \cmidrule(lr){7-8} \cmidrule(lr){9-10} \cmidrule(lr){11-12}
\textbf{Method} & \textbf{Metric} & S1 & S2 & S1 & S2 & S1 & S2 & S1 & S2 & S1 & S2 \\
\midrule
\multirow{3}{*}{$\pi_{0.5}$}
& CAR ($\uparrow$)  & 17.00 & 13.50 & 7.50 & 40.00 & 12.50 & 33.50 & 12.50 & 13.00 & 12.40 & 25.00 \\
& TSR ($\uparrow$)  & 60.00 & 59.50 & 45.50 & 63.00 & 36.50 & 71.00 & 42.00 & 29.50 & 46.00 & 55.80 \\
& ETS ($\downarrow$)  & 199.80 & 203.50 & 236.00 & 184.60 & 255.50 & \textbf{190.50} & \textbf{459.40} & 496.60 & 287.70 & 268.80 \\
\cmidrule{1-12}
\multirow{3}{*}{AEGIS}
& CAR ($\uparrow$)  & 73.50 & \textbf{77.50} & 92.00 & 71.00 & 58.50 & \textbf{91.00} & 78.00 & 81.30 & 75.50 & 80.20 \\
& TSR ($\uparrow$)  & 65.50 & \textbf{81.00} & 74.50 & 76.00 & 84.00 & 76.50 & 51.00 & 36.50 & 68.80 & 67.50 \\
& ETS ($\downarrow$)  & 202.90 & \textbf{173.60} & 190.90 & \textbf{168.30} & \textbf{206.30} & 196.30 & 464.20 & \textbf{496.10} & \textbf{266.10} & \textbf{258.60} \\
\cmidrule{1-12}
\multirow{3}{*}{\textbf{Ours}}
& CAR ($\uparrow$)  & \textbf{77.50} & 75.50 & \textbf{94.00} & \textbf{82.50} & \textbf{82.50} & 85.50 & \textbf{82.00} & \textbf{83.00} & \textbf{84.00} & \textbf{81.62} \\
& TSR ($\uparrow$)  & \textbf{75.50} & 78.00 & \textbf{91.00} & \textbf{83.50} & \textbf{90.50} & \textbf{81.00} & \textbf{81.50} & \textbf{72.00} & \textbf{84.62} & \textbf{78.62} \\
& ETS ($\downarrow$)  & \textbf{192.62} & 180.08 & \textbf{176.80} & 238.30 & 269.45 & 341.80 & 464.60 & 536.10 & 275.87 & 324.07 \\
\bottomrule
\end{tabular}}
\end{table}

\subsection{Per-Task Analysis}

Per-task results across all suites and safety levels are reported in Tables \ref{tab:expanded-results-A} and  \ref{tab:expanded-results-B} in Appendix~\ref{app:additional_results}. Several patterns merit discussion.

\paragraph{Goal suite.} Our method achieves its strongest results on Goal tasks, reaching 100\% TSR on Goal Task 1 at Safety Level I (vs.\ 80\% for AEGIS) and 98\% TSR on Goal Task 2 at Level II. These tasks involve placing a bowl at a specified location, where the primary challenge is navigating around an obstacle en route. The predictive nature of our guidance is well-suited here: the trajectory-level view identifies the obstacle early and distributes small corrections across the approach, rather than requiring a sharp correction near the obstacle.

\paragraph{Long-horizon tasks.} The Long suite shows the largest gap between our method and AEGIS. On Long Task 1 (placing alphabet soup and cream cheese in a basket), our method achieves 84\% and 90\% TSR at Levels I and II respectively, compared to 76\% and 48\% for AEGIS. Multi-step tasks require the robot to return near obstacles repeatedly; post-hoc corrections at each encounter compound, progressively pushing the policy further from its learned distribution. In-generation guidance mitigates this accumulation by allowing the model to incorporate corrections into its trajectory planning at each action chunk.

\paragraph{Challenging cases and limitations.} Performance is weakest on Spatial Task 1 (picking up a bowl between a ramekin and plate), where our TSR of 70\% at Level II trails AEGIS, as the obstacle forces the arm's joints to bend far beyond their typical configuration. This is a consequence of two facts: our barriers act only on the end-effector position under single-integrator dynamics, so collisions with the rest of the arm's body are unmodeled and generalizing to higher-order or full-body dynamics is left to future work; and instantiating the barriers requires known obstacle geometry and positions, which we take from the simulator. The latter isolates the control contribution from perception but departs from a central motivation for VLAs---acting from raw observations without an explicit state---so deriving constraints from perception is an important next step. Object Task 3 at Level I also shows lower CAR (46\% vs.\ AEGIS' 78\%), suggesting the ellipsoidal approximation can underestimate required clearance for some geometries. Finally, our ETS exceeds other methods on three of four suites, most notably on Object; these delays stem mainly from the policy emitting neutral actions under uninformative sensory inputs rather than from the guidance process itself.

\section{Conclusion}
\label{sec:conclusion}

We presented a neuro-symbolic safety guidance mechanism for Vision-Language-Action models that integrates barrier constraint satisfaction directly into the flow matching action generation process. By predicting trajectories from intermediate denoising states and applying minimum-norm corrections when collisions are anticipated, our approach steers action chunks away from obstacles while allowing subsequent denoising steps to adapt to corrections. Evaluation on SafeLIBERO demonstrates that this predictive, in-generation approach achieves 82.81\% collision avoidance and 81.62\% task success, improving over both the unguided baseline and the single-action CBF approach. The largest gains appear on long-horizon tasks, where the compounding effects of post-hoc distribution shift are most pronounced. These results suggest that integrating safety constraints into the generative process offers a promising direction for deploying VLA models in safety-critical environments.

\bibliography{main}

\clearpage
\appendix

\section{Algorithm and Parameters}
\label{app:alg}
\begin{algorithm}[!h]
\caption{Predictive Safety-Guided Flow Matching}
\label{alg:guided_flow}
\begin{algorithmic}[1]
\REQUIRE Obstacles $\mathcal{O} = \{O_k\}_{k=1}^{K}$, semi-axes $r=(r_x,r_y,r_z)$, margin $d_{\text{safe}}$,
         decay rate $\gamma$, steps $N$, horizon $H$, velocity field $v_\theta$
\ENSURE Safe executed trajectory $\{s_0, s_1, \ldots\}$
\WHILE{task not completed}
    \STATE Get observation $o$ and end-effector position $p_{\text{ee}}^{(0)}$
    \STATE Sample $A^\tau \sim \mathcal{N}(0, I)$; \quad $\tau \gets 1$; \quad $\Delta\tau \gets -1/N$
    \WHILE{$\tau > 0$}
        \STATE $v \gets v_\theta(A^\tau, \tau \mid o)$ \hfill $\triangleright$ Model velocity
        \STATE $A^{\tau + \Delta\tau} \gets A^\tau + v \cdot \Delta\tau$ \hfill $\triangleright$ Euler step
        \STATE Predict $\{p_{\text{ee}}^{(j)}\}_{j=0}^{H}$ from $A^{\tau + \Delta\tau}$ via Eq.~\eqref{eq:trajectory_prediction}
        \STATE Evaluate $B_j$, $j = 0,\ldots,H$ via Eqs.~\eqref{eq:signed_distance}--\eqref{eq:barrier}
        \IF{$\exists\, j : B_j < (1-\gamma)\, B_{j-1}$} 
            \STATE Solve Eq.~\eqref{eq:cbf_qp} for min-norm correction ${\delta}^*$ \hfill $\triangleright$ Trajectory-level CBF violation
            \STATE $A^{\tau + \Delta\tau} \gets A^{\tau + \Delta\tau} + \delta^*$
        \ENDIF
        \STATE $\tau \gets \tau + \Delta\tau$
    \ENDWHILE
    \STATE Execute safe chunk $A^0$ on the robot; update $p_{\text{ee}}$ and states
\ENDWHILE
\end{algorithmic}
\end{algorithm}

\begin{table}[ht]
\centering
\small
\begin{tabular}{lll}
\toprule
Parameter & Symbol & Value \\
\midrule
End-effector semi-axes & $(r_x, r_y, r_z)$ & $(0.06, 0.12, 0.11)$ m \\
Safety margin & $d_{\text{safe}}$ & 0.01 m \\
CBF decay rate & $\gamma$ & 0.9 \\
\bottomrule
\end{tabular}
\caption{Safety guidance parameters.}
\label{tab:hyperparams}
\end{table}

\clearpage
\section{SafeLIBERO Task Details}
\label{app:safelibero}

\begin{figure*}[!h]
\vspace{-20.0pt}
    \centering
    \setlength{\tabcolsep}{1pt}
    \renewcommand{\arraystretch}{0.5}
    \begin{tabular}{cccc}
        \multicolumn{4}{c}{\textbf{Spatial}} \\[2pt]
        \includegraphics[width=0.24\textwidth]{figures/safelibero_tasks/safelibero_spatial_I/task_id_0_pick_up_the_black_bowl_on_the_wooden_cabinet_and_place_it_on_the_plate.png} & \includegraphics[width=0.24\textwidth]{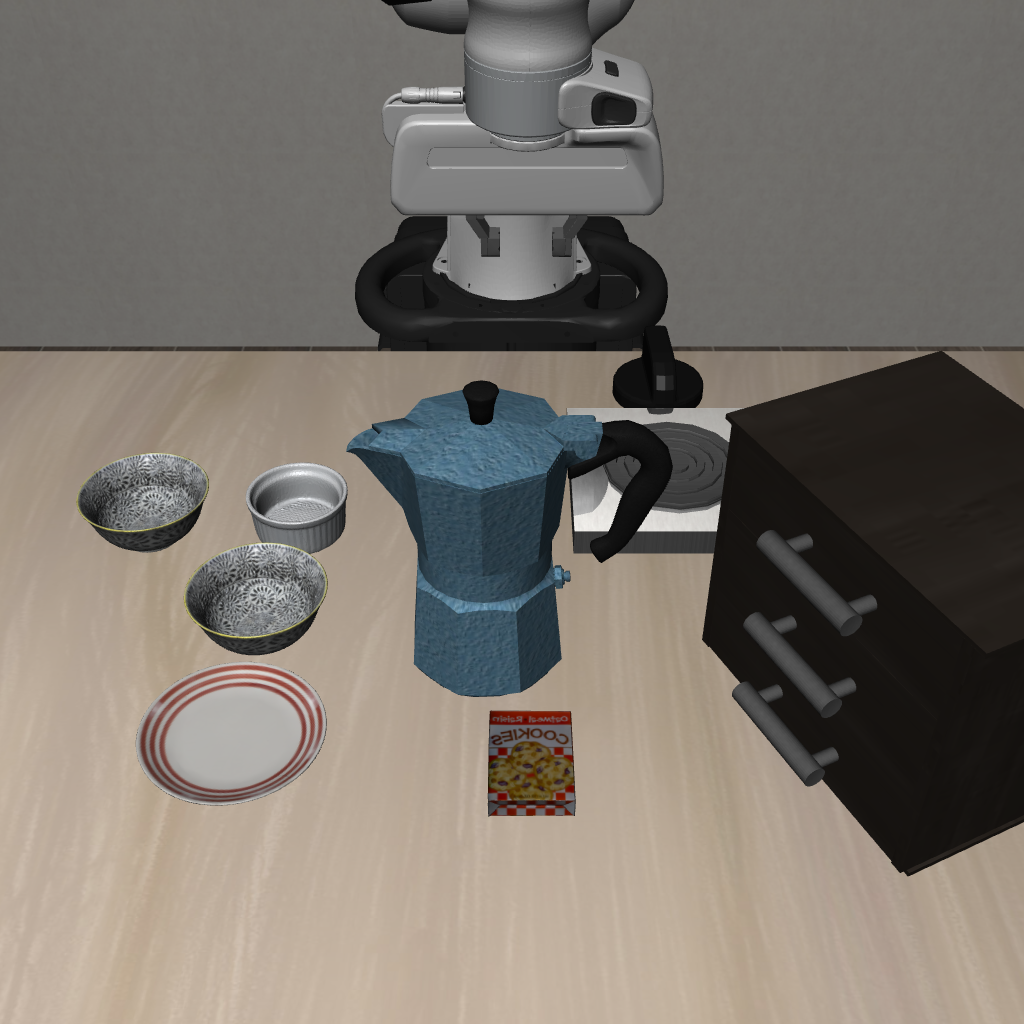} & \includegraphics[width=0.24\textwidth]{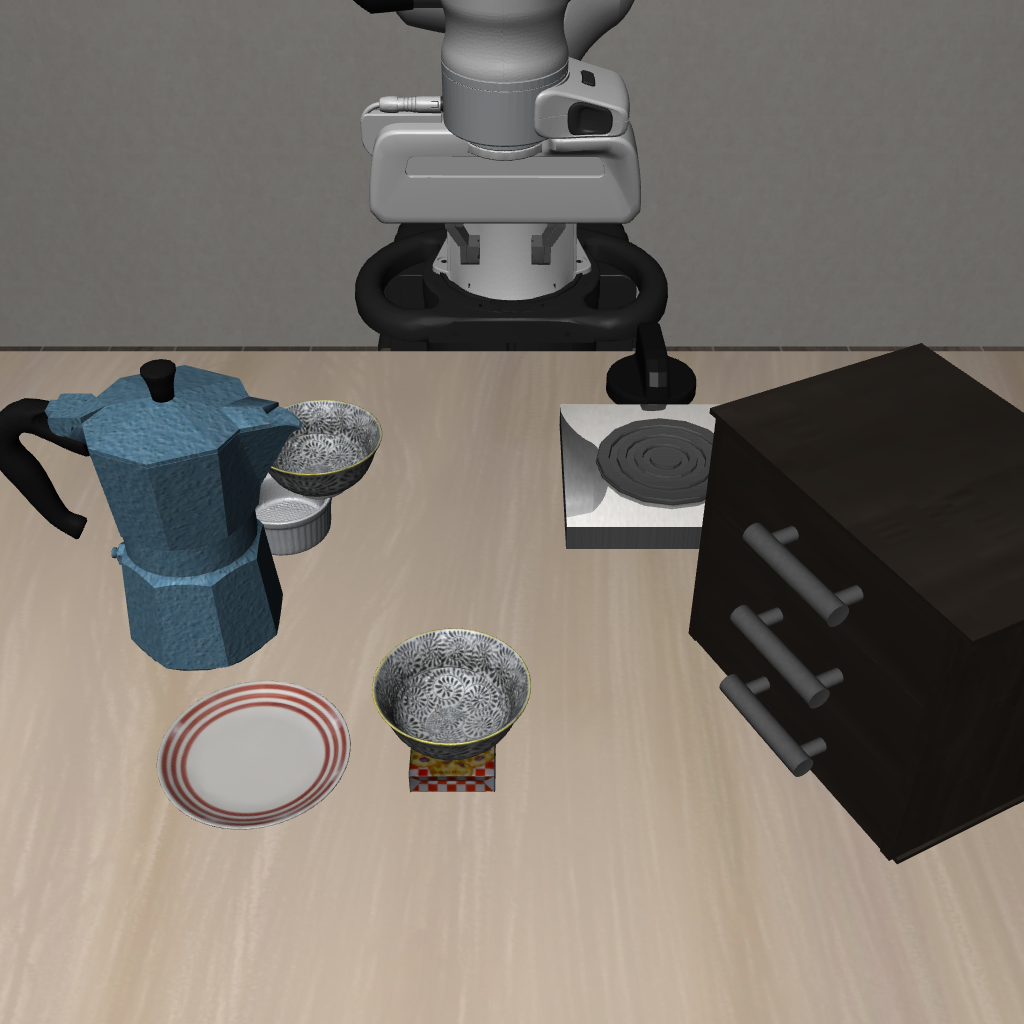} & \includegraphics[width=0.24\textwidth]{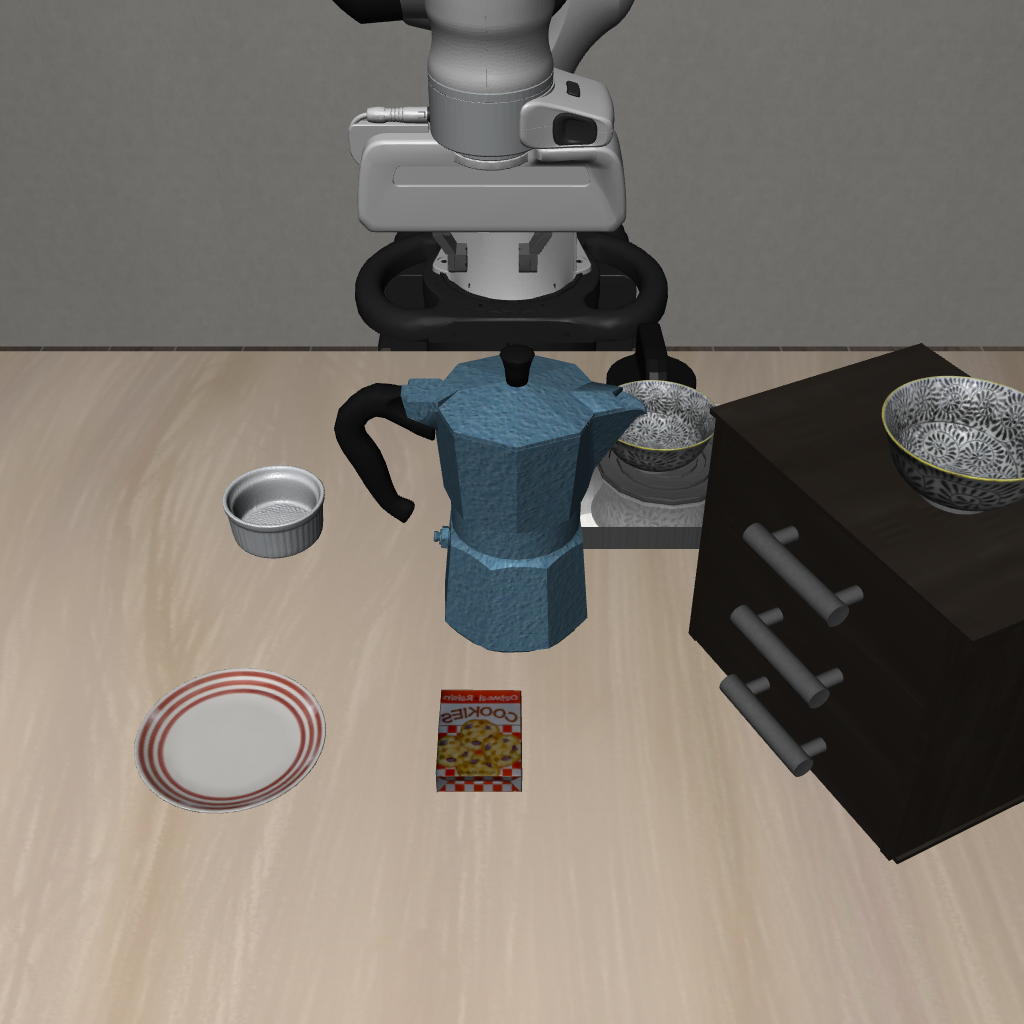} \\[1pt]
        \parbox[t]{0.24\textwidth}{\centering\scriptsize Pick up the black bowl on the wooden cabinet and place it on the plate} & \parbox[t]{0.24\textwidth}{\centering\scriptsize Pick up the black bowl between the plate and the ramekin and place it on the plate} & \parbox[t]{0.24\textwidth}{\centering\scriptsize Pick up the black bowl on the ramekin and place it on the plate} & \parbox[t]{0.24\textwidth}{\centering\scriptsize Pick up the black bowl on the stove and place it on the plate} \\[6pt]
        \multicolumn{4}{c}{\textbf{Object}} \\[2pt]
        \includegraphics[width=0.24\textwidth]{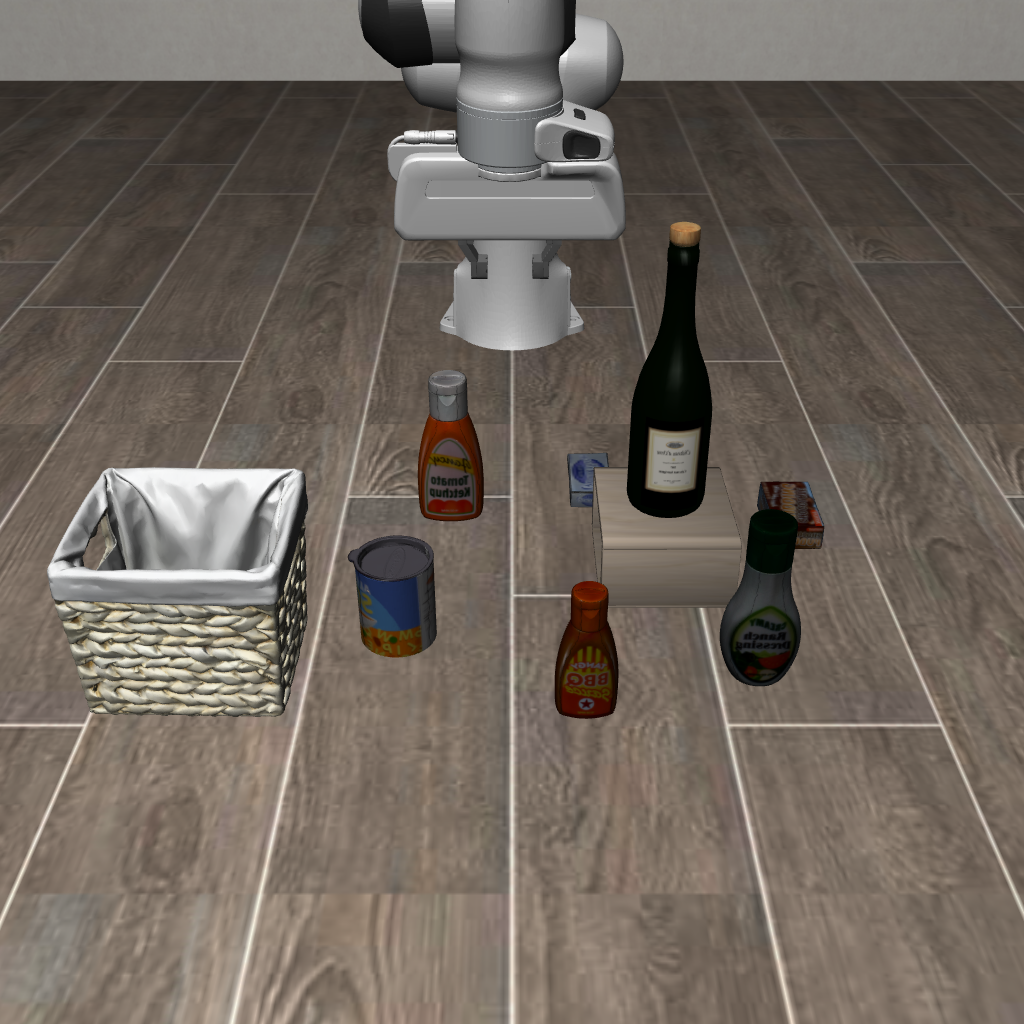} & \includegraphics[width=0.24\textwidth]{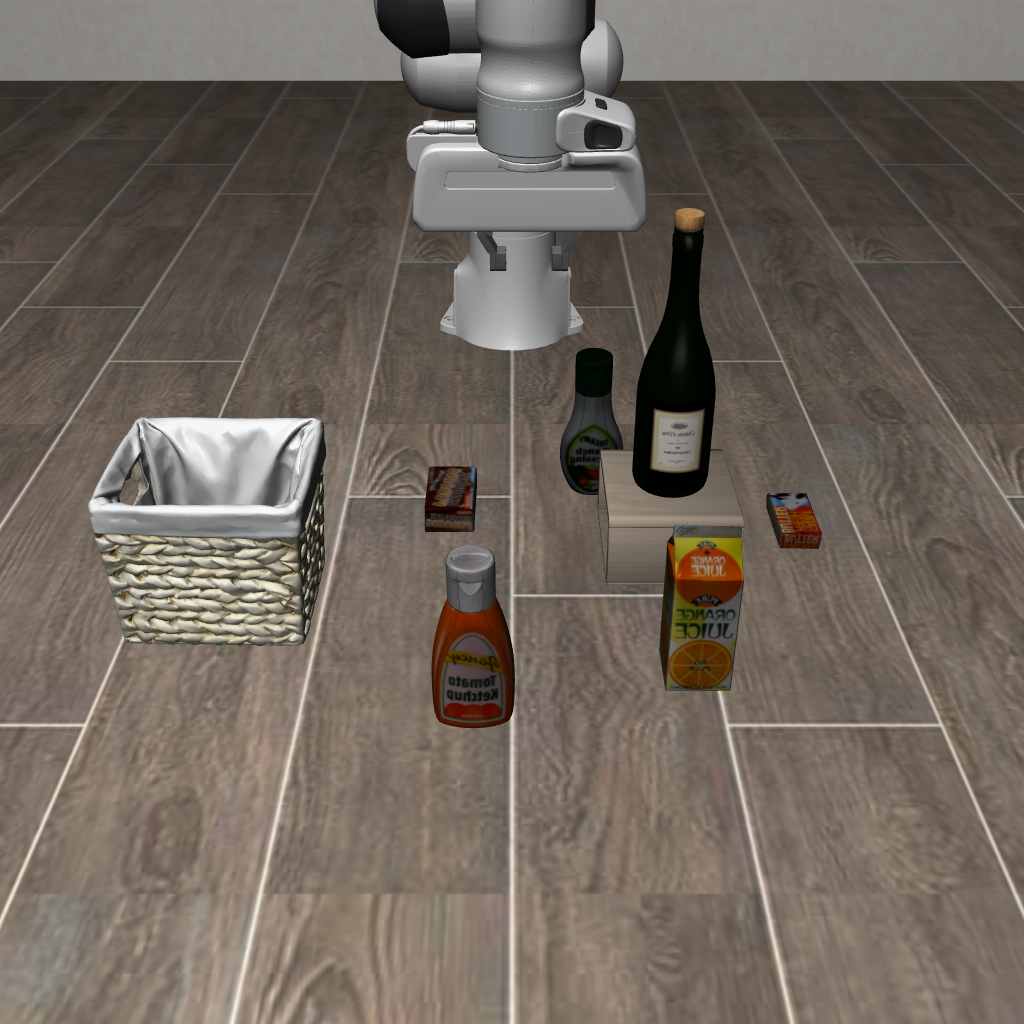} & \includegraphics[width=0.24\textwidth]{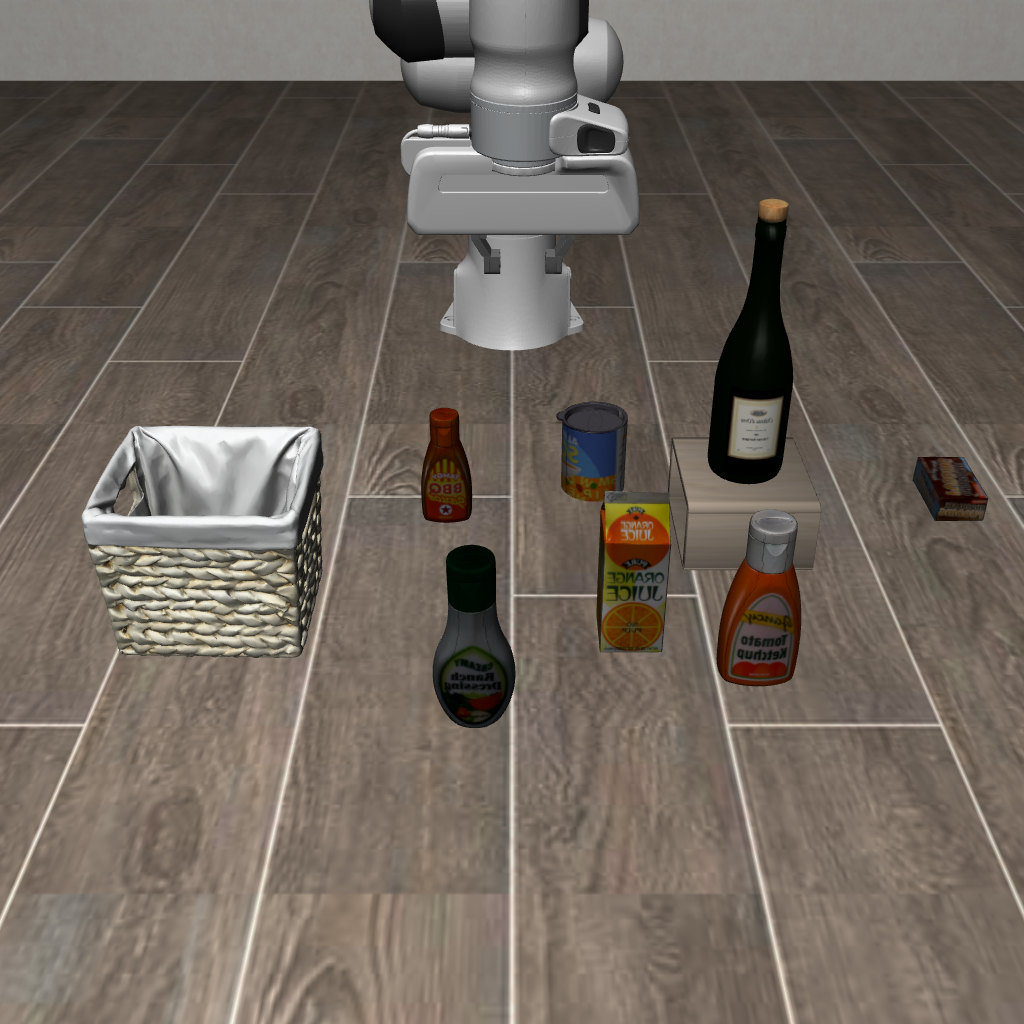} & \includegraphics[width=0.24\textwidth]{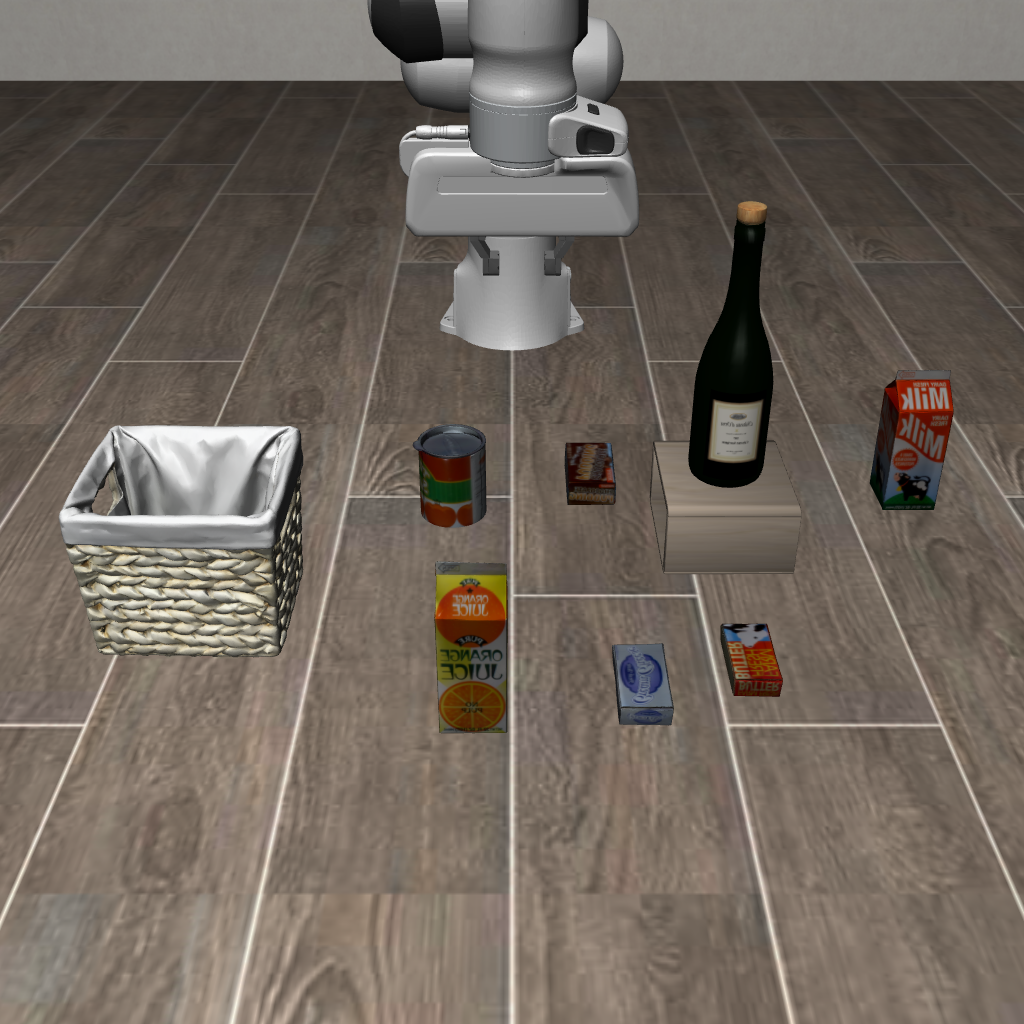} \\[1pt]
        \parbox[t]{0.24\textwidth}{\centering\scriptsize Pick up the bbq sauce and place it in the basket} & \parbox[t]{0.24\textwidth}{\centering\scriptsize Pick up the orange juice and place it in the basket} & \parbox[t]{0.24\textwidth}{\centering\scriptsize Pick up the chocolate pudding and place it in the basket} & \parbox[t]{0.24\textwidth}{\centering\scriptsize Pick up the milk and place it in the basket} \\[6pt]
        \multicolumn{4}{c}{\textbf{Goal}} \\[2pt]
        \includegraphics[width=0.24\textwidth]{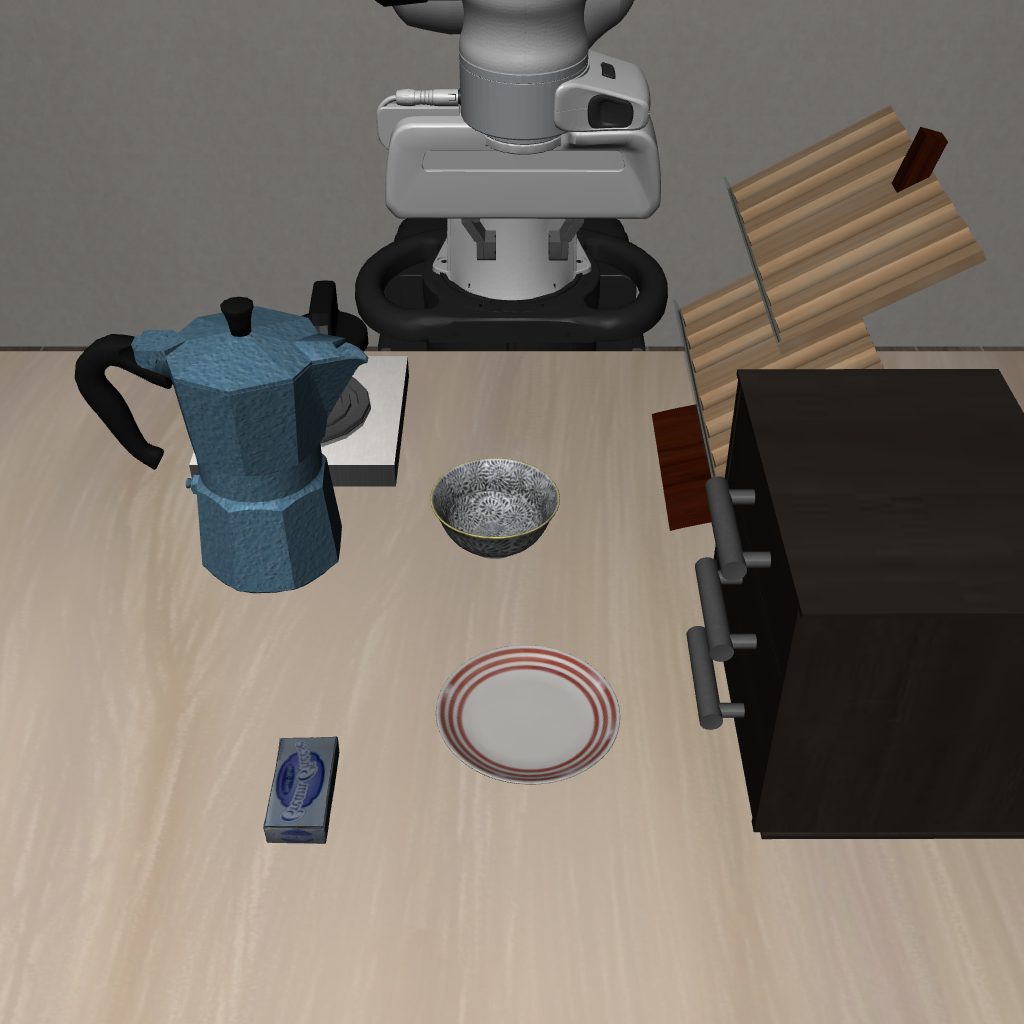} & \includegraphics[width=0.24\textwidth]{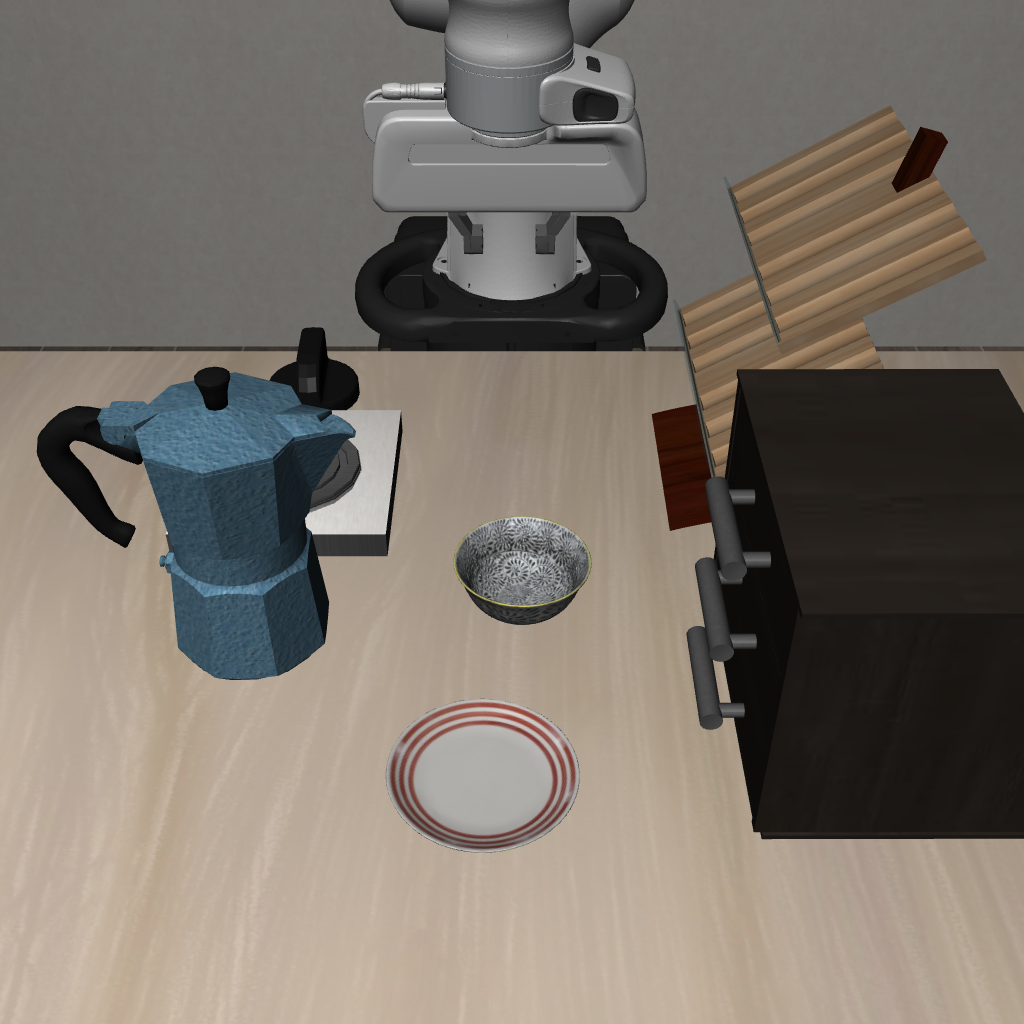} & \includegraphics[width=0.24\textwidth]{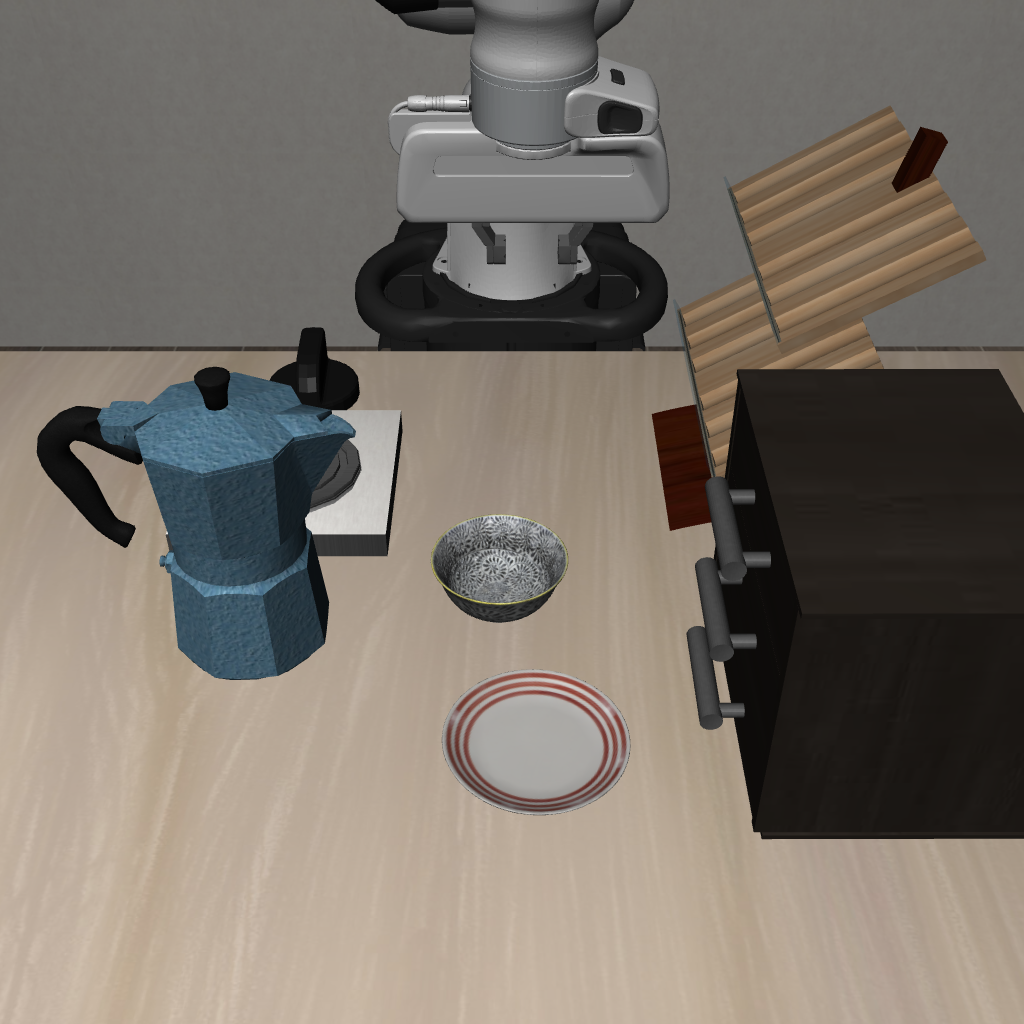} & \includegraphics[width=0.24\textwidth]{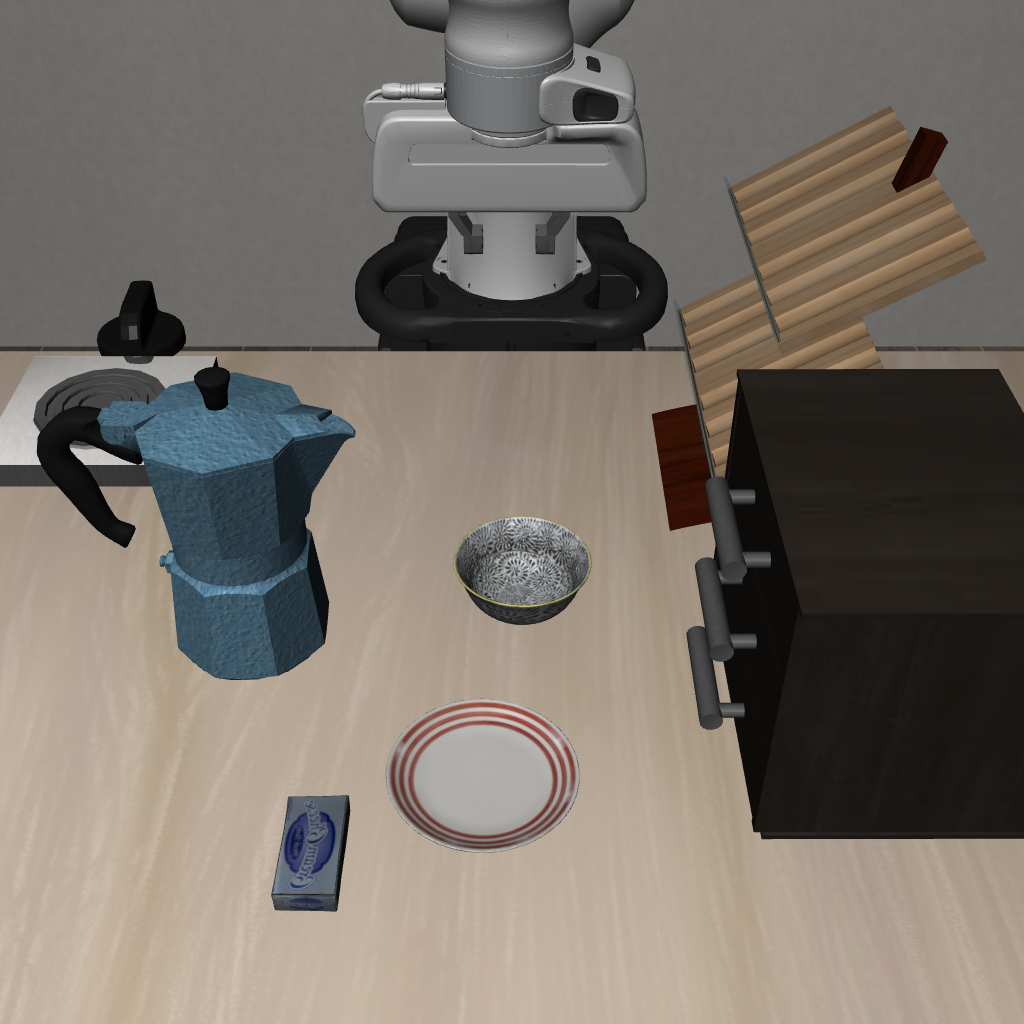} \\[1pt]
        \parbox[t]{0.24\textwidth}{\centering\scriptsize Open the top drawer and put the bowl inside} & \parbox[t]{0.24\textwidth}{\centering\scriptsize Put the bowl on the plate} & \parbox[t]{0.24\textwidth}{\centering\scriptsize Put the bowl on top of the cabinet} & \parbox[t]{0.24\textwidth}{\centering\scriptsize Put the bowl on the stove} \\[6pt]
        \multicolumn{4}{c}{\textbf{Long}} \\[2pt]
        \includegraphics[width=0.24\textwidth]{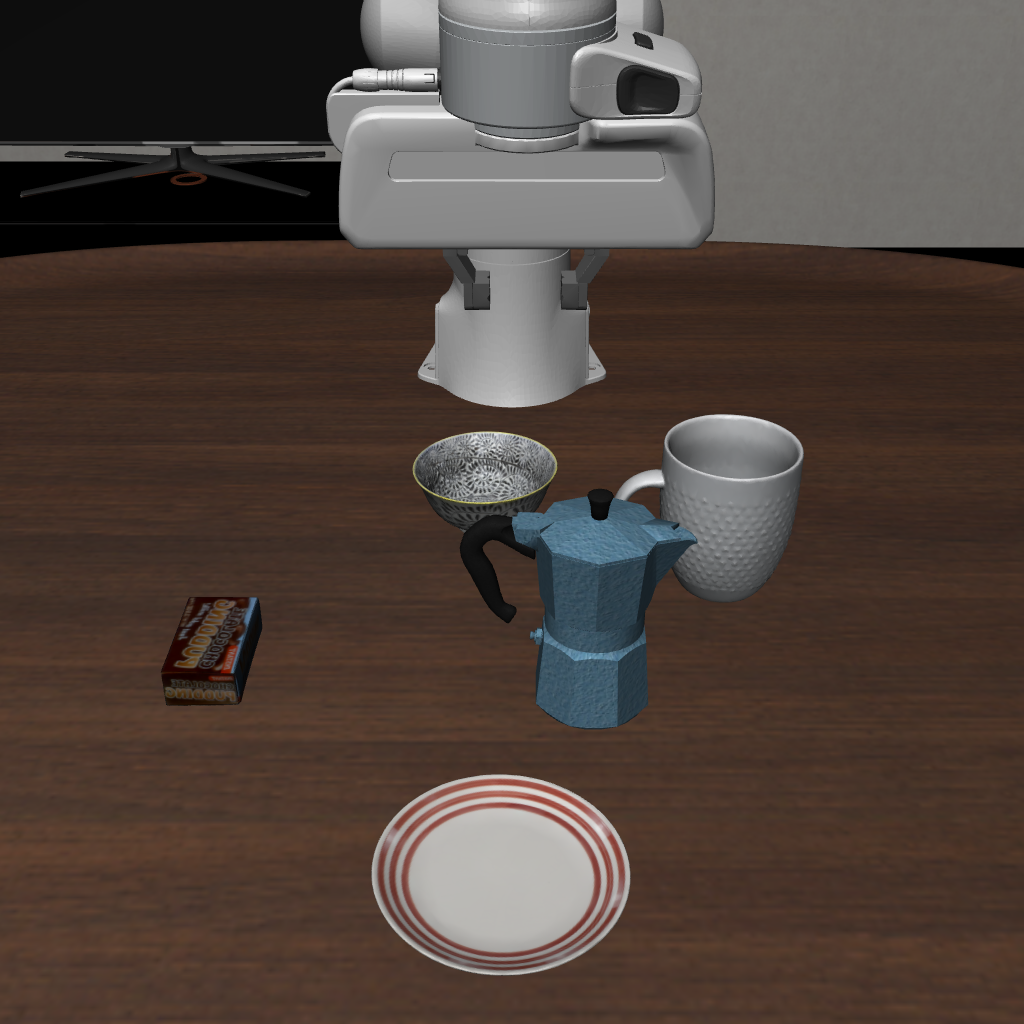} & \includegraphics[width=0.24\textwidth]{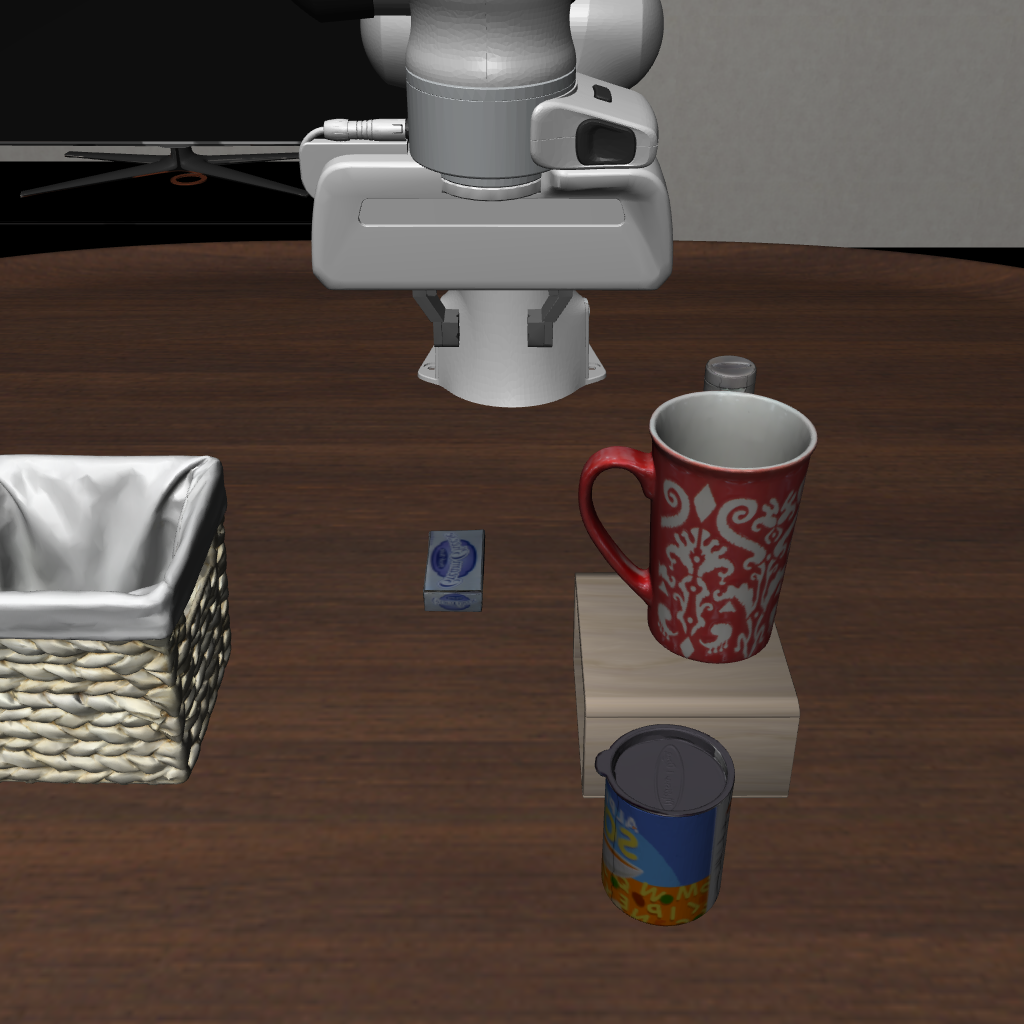} & \includegraphics[width=0.24\textwidth]{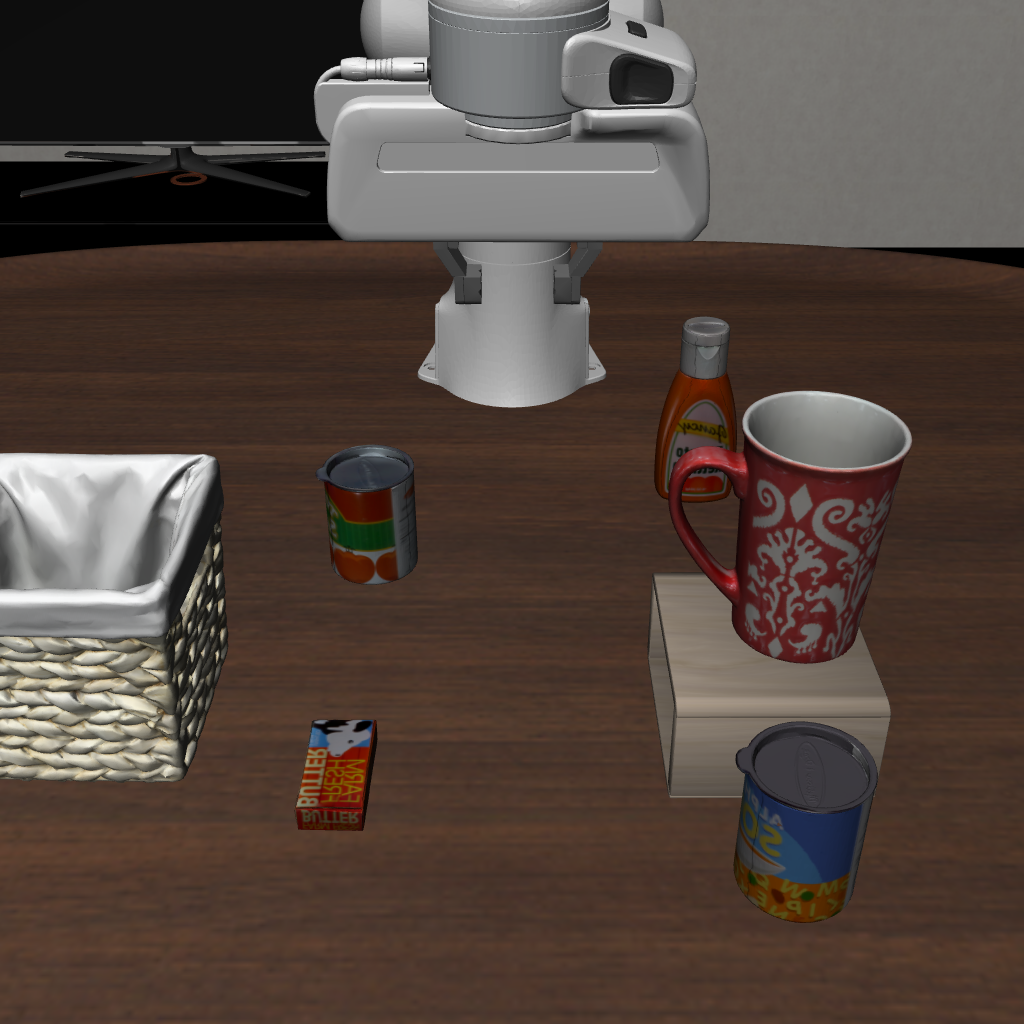} & \includegraphics[width=0.24\textwidth]{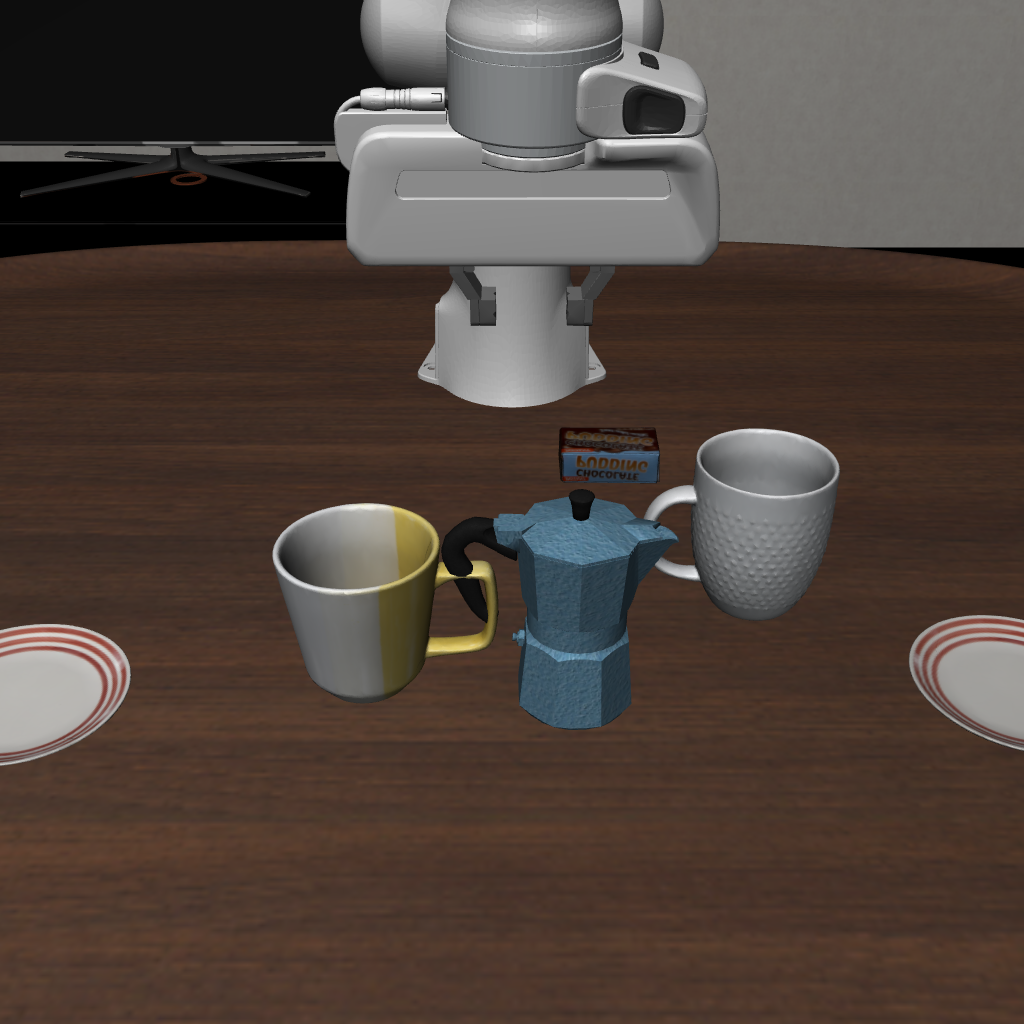} \\[1pt]
        \parbox[t]{0.24\textwidth}{\centering\scriptsize Put the white mug on the plate and put the chocolate pudding to the right of the plate} & \parbox[t]{0.24\textwidth}{\centering\scriptsize Put both the alphabet soup and the cream cheese box in the basket} & \parbox[t]{0.24\textwidth}{\centering\scriptsize Put both the alphabet soup and the tomato sauce in the basket} & \parbox[t]{0.24\textwidth}{\centering\scriptsize Put the white mug on the left plate and put the yellow and white mug on the right plate} \\
    \end{tabular}
    \caption{SafeLIBERO Safety Level I tasks. Four suites (Spatial, Object, Goal, Long) with obstacles at safety level i density.}
    \label{fig:safelibero_tasks_i}
\end{figure*}

\begin{figure*}[!h]
    \centering
    \setlength{\tabcolsep}{1pt}
    \renewcommand{\arraystretch}{0.5}
    \begin{tabular}{cccc}
        \multicolumn{4}{c}{\textbf{Spatial}} \\[2pt]
        \includegraphics[width=0.24\textwidth]{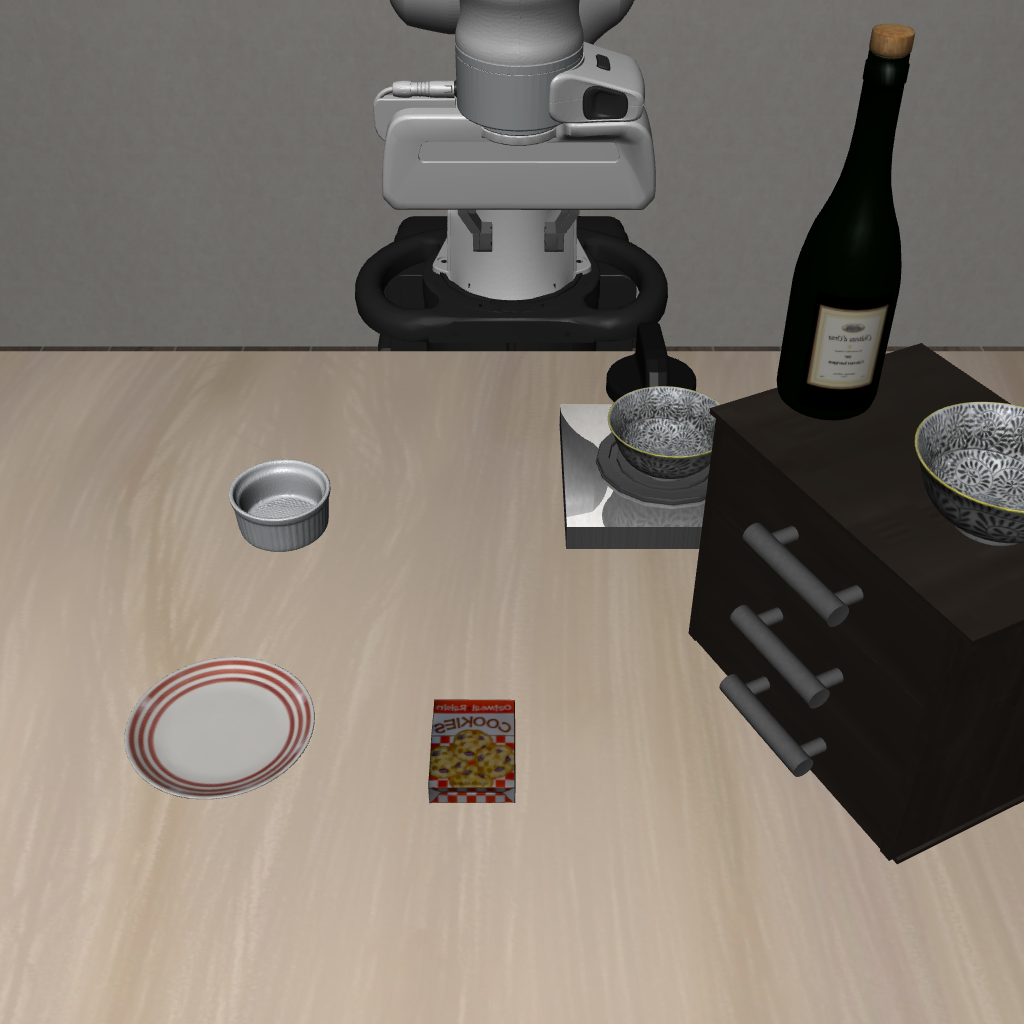} & \includegraphics[width=0.24\textwidth]{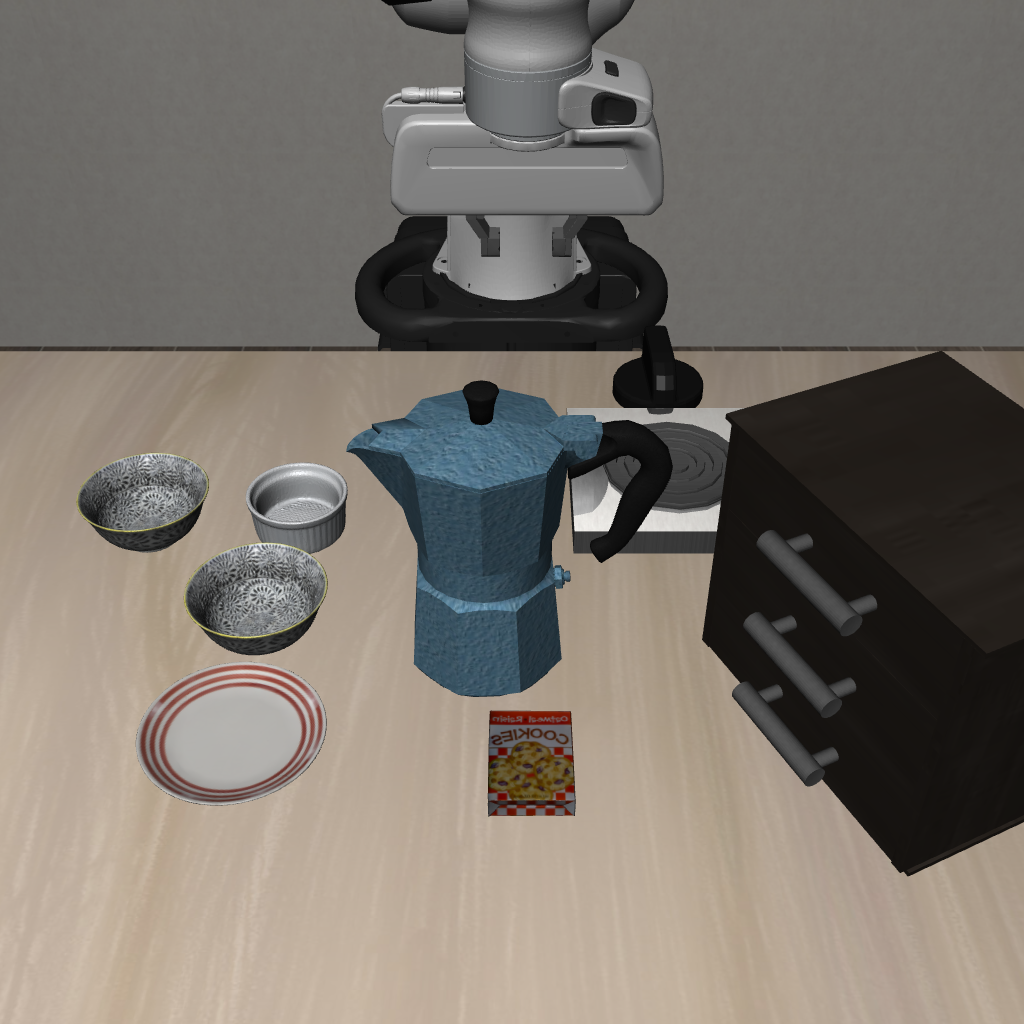} & \includegraphics[width=0.24\textwidth]{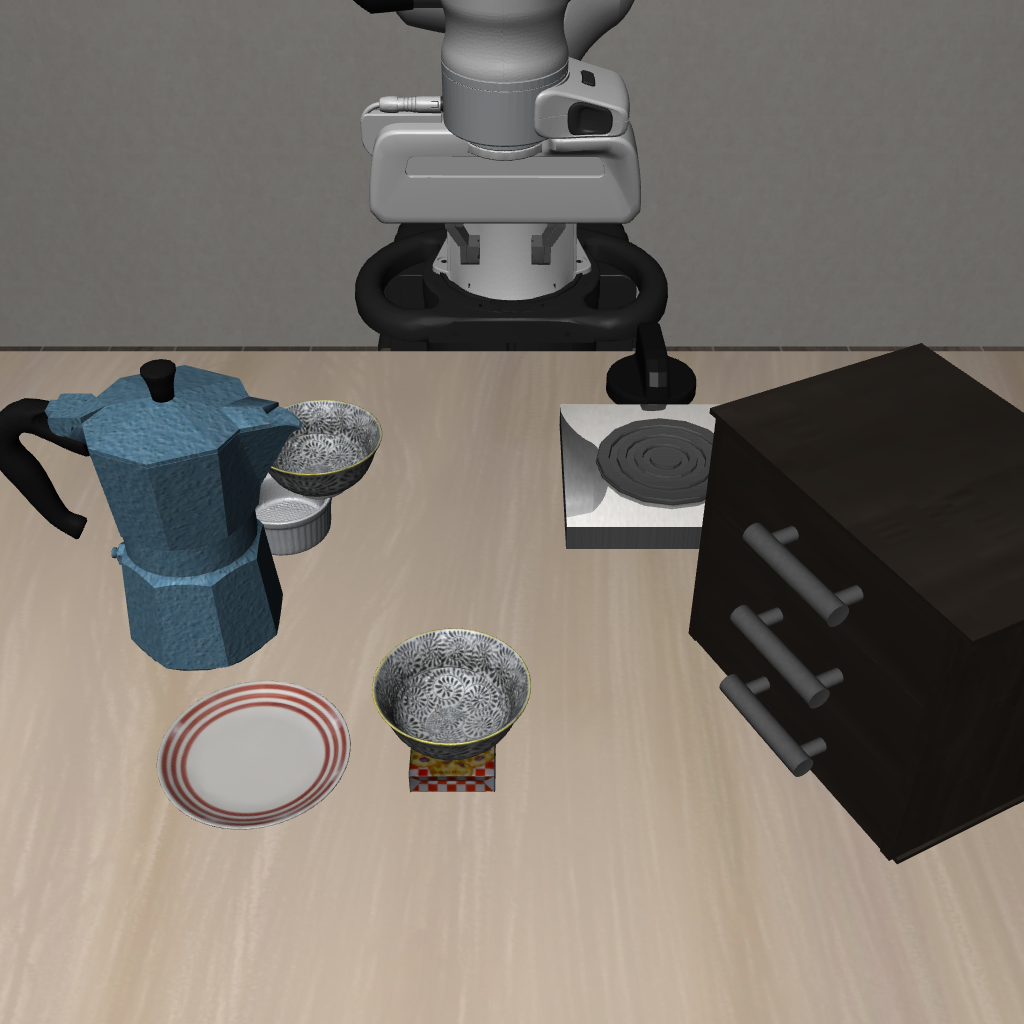} & \includegraphics[width=0.24\textwidth]{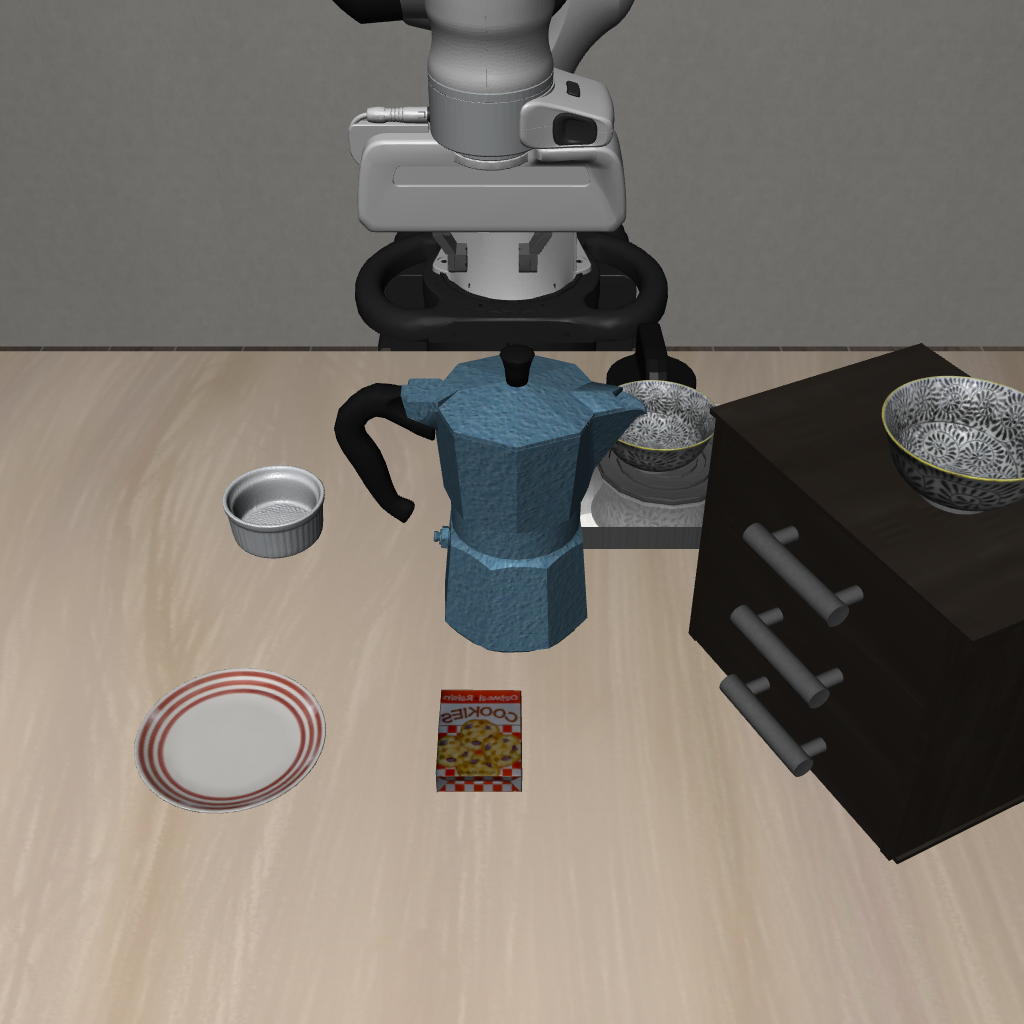} \\[1pt]
        \parbox[t]{0.24\textwidth}{\centering\scriptsize Pick up the black bowl on the wooden cabinet and place it on the plate} & \parbox[t]{0.24\textwidth}{\centering\scriptsize Pick up the black bowl between the plate and the ramekin and place it on the plate} & \parbox[t]{0.24\textwidth}{\centering\scriptsize Pick up the black bowl on the ramekin and place it on the plate} & \parbox[t]{0.24\textwidth}{\centering\scriptsize Pick up the black bowl on the stove and place it on the plate} \\[6pt]
        \multicolumn{4}{c}{\textbf{Object}} \\[2pt]
        \includegraphics[width=0.24\textwidth]{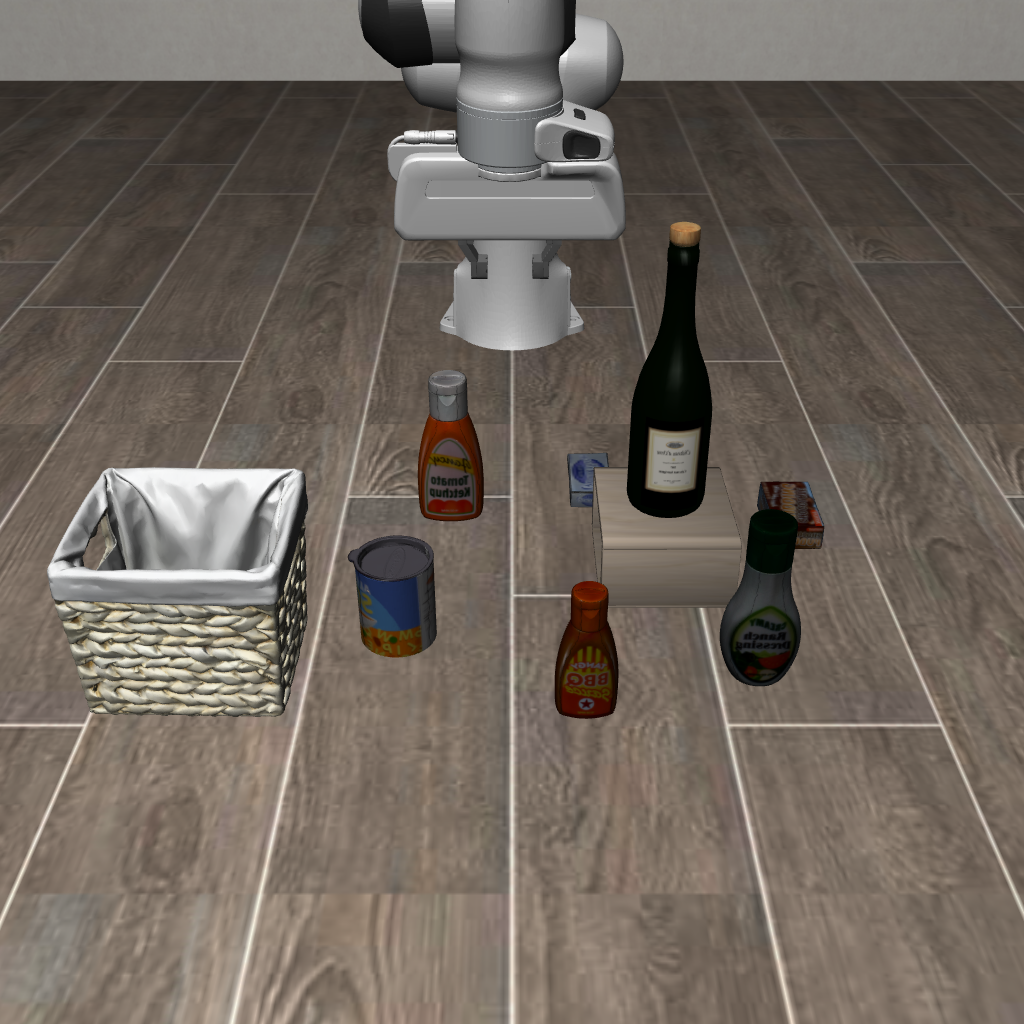} & \includegraphics[width=0.24\textwidth]{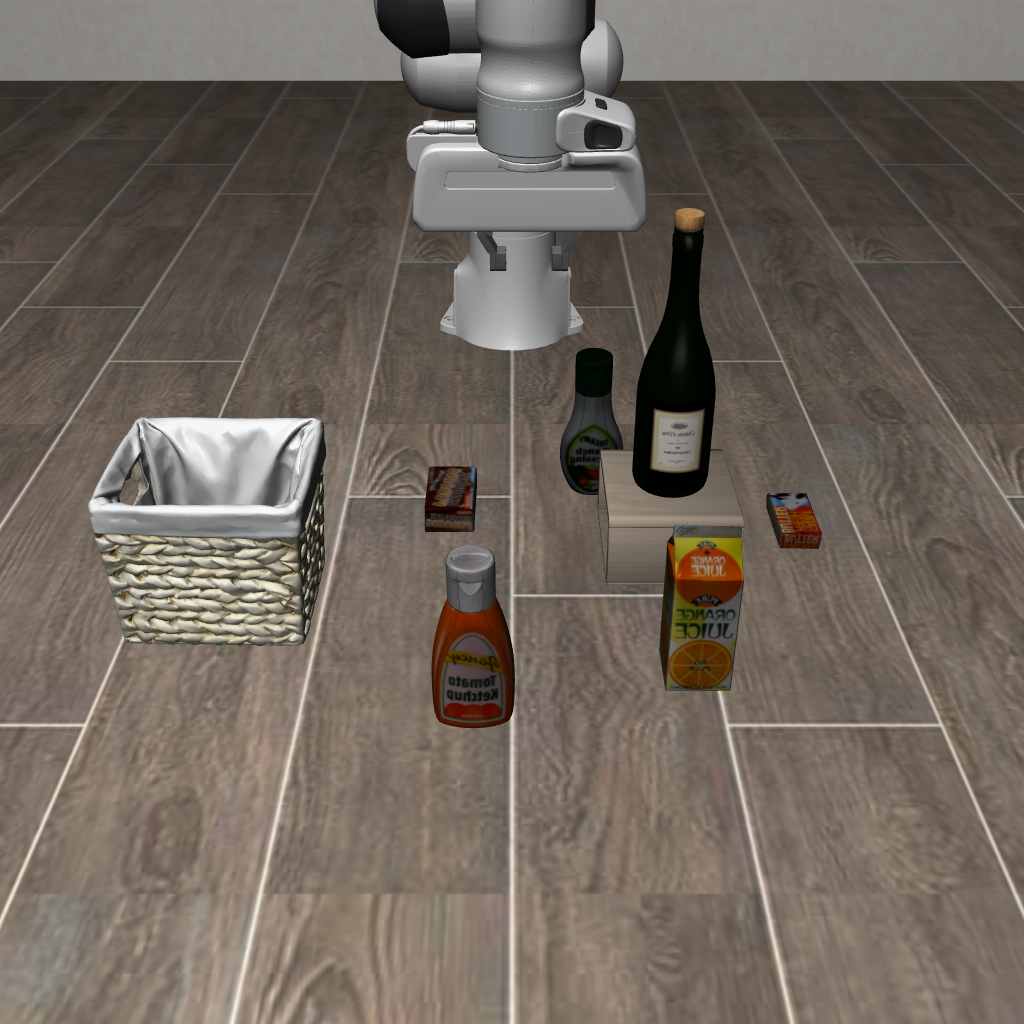} & \includegraphics[width=0.24\textwidth]{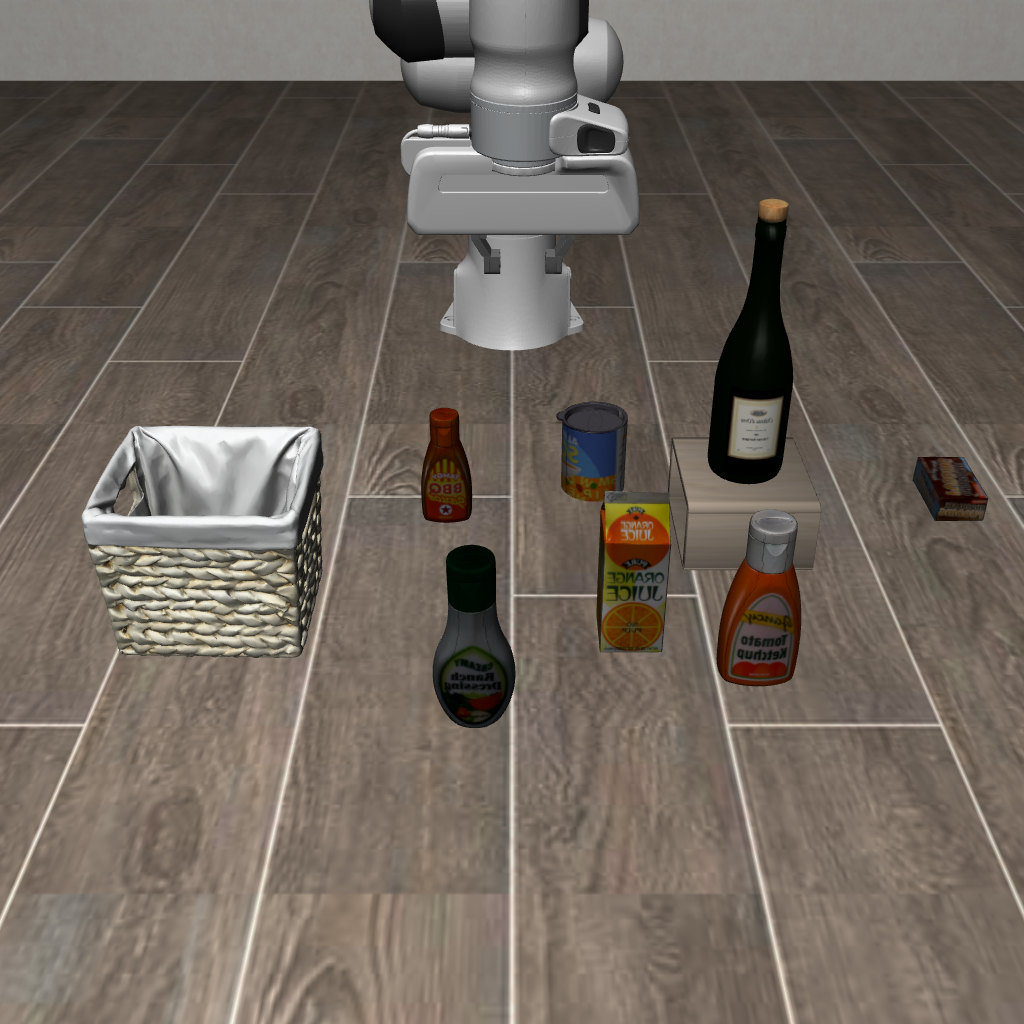} & \includegraphics[width=0.24\textwidth]{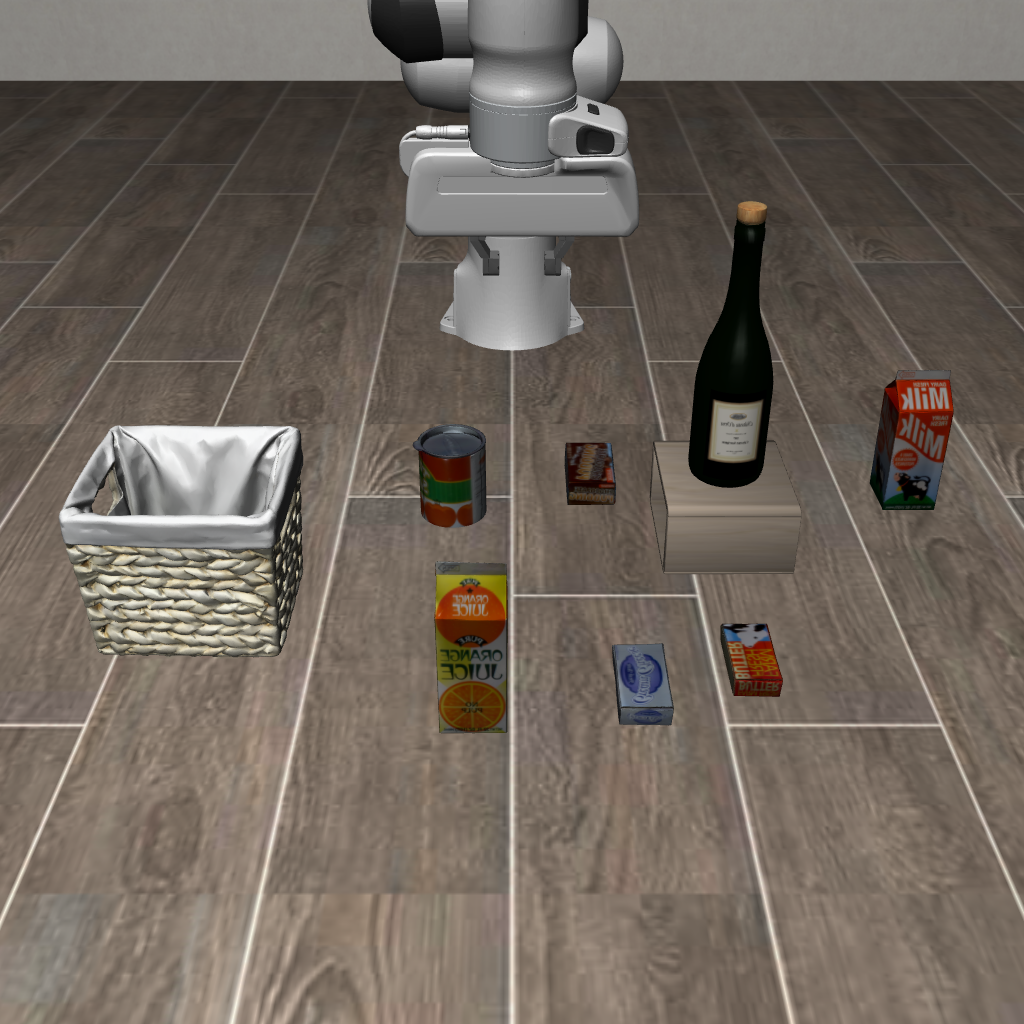} \\[1pt]
        \parbox[t]{0.24\textwidth}{\centering\scriptsize Pick up the bbq sauce and place it in the basket} & \parbox[t]{0.24\textwidth}{\centering\scriptsize Pick up the orange juice and place it in the basket} & \parbox[t]{0.24\textwidth}{\centering\scriptsize Pick up the chocolate pudding and place it in the basket} & \parbox[t]{0.24\textwidth}{\centering\scriptsize Pick up the milk and place it in the basket} \\[6pt]
        \multicolumn{4}{c}{\textbf{Goal}} \\[2pt]
        \includegraphics[width=0.24\textwidth]{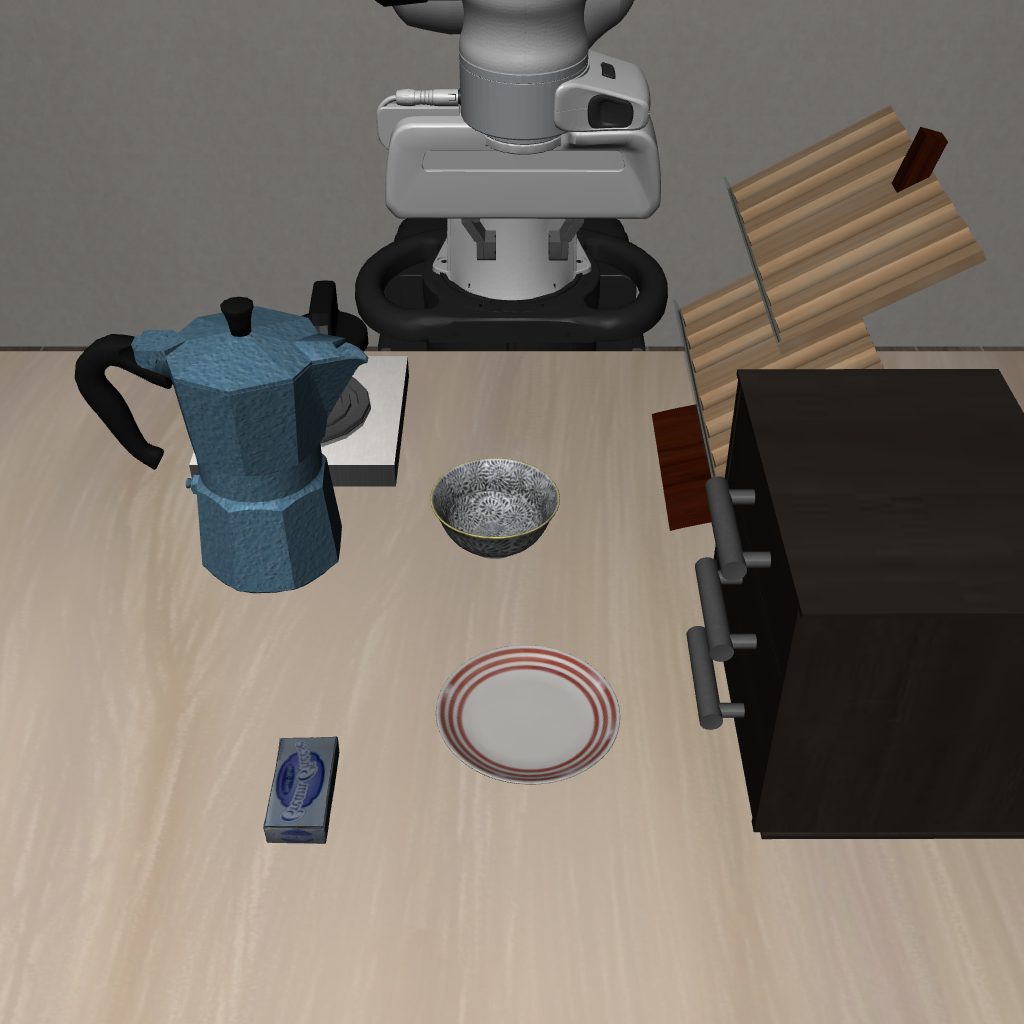} & \includegraphics[width=0.24\textwidth]{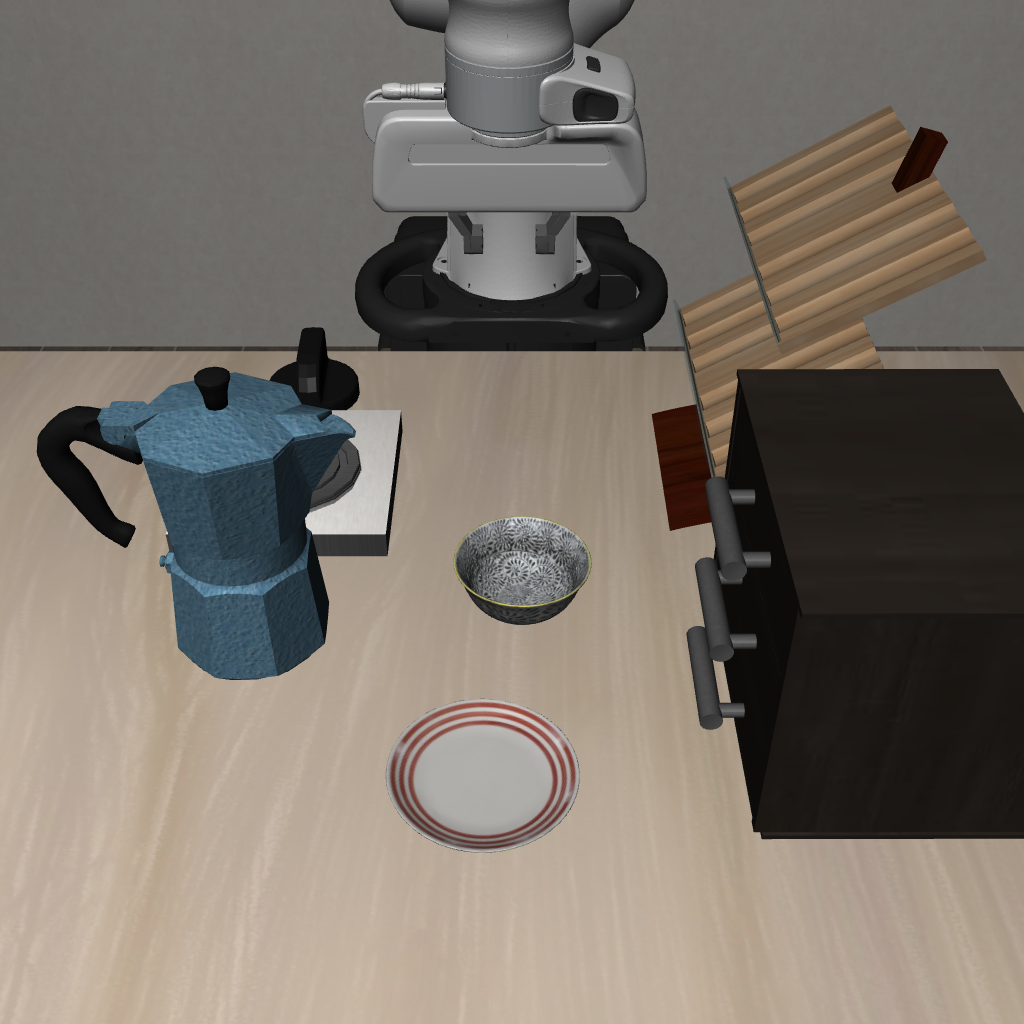} & \includegraphics[width=0.24\textwidth]{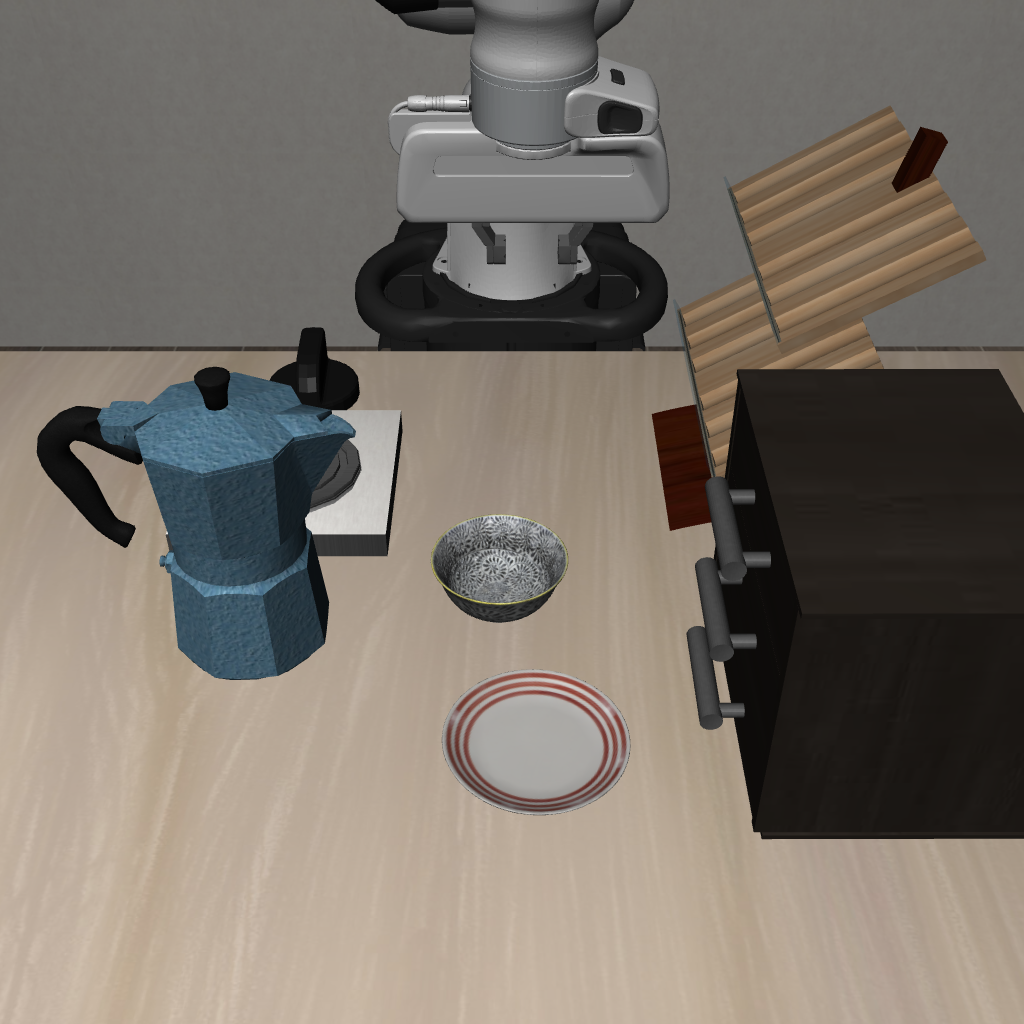} & \includegraphics[width=0.24\textwidth]{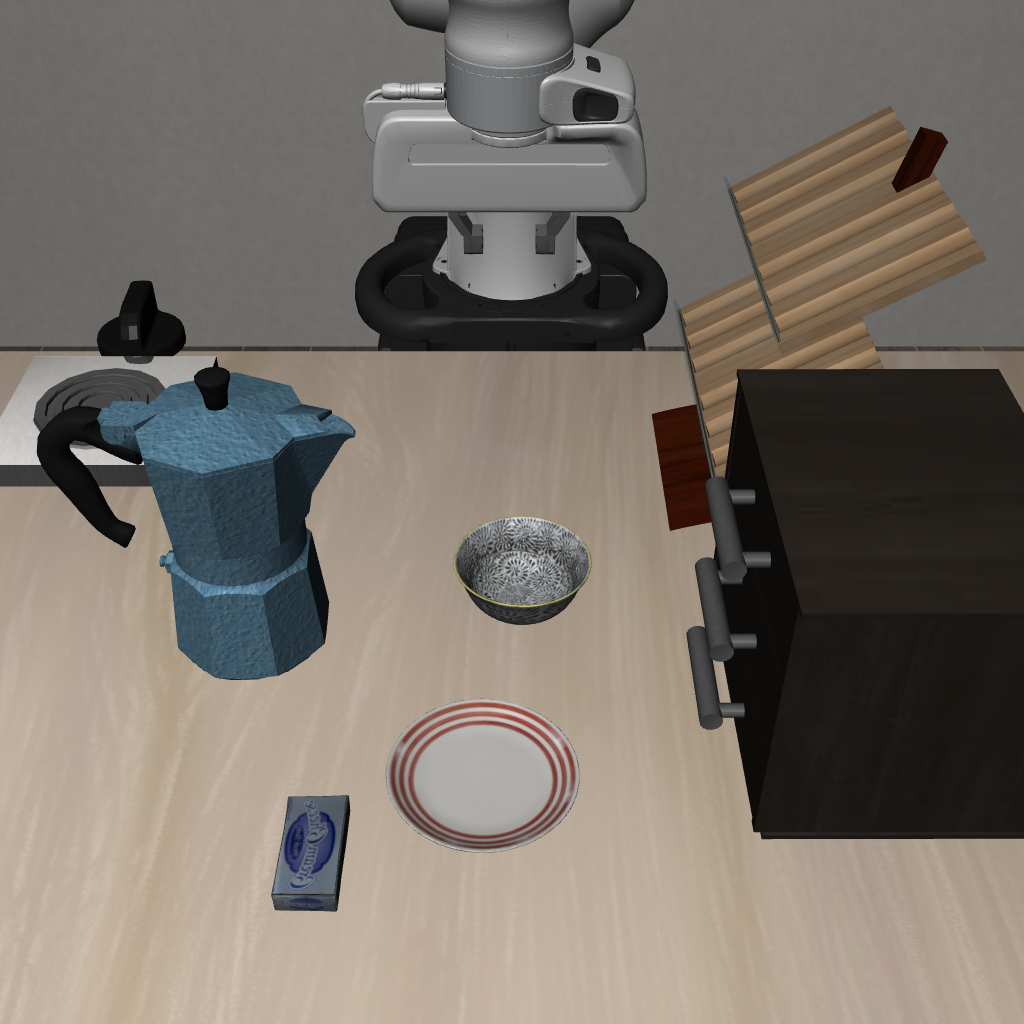} \\[1pt]
        \parbox[t]{0.24\textwidth}{\centering\scriptsize Open the top drawer and put the bowl inside} & \parbox[t]{0.24\textwidth}{\centering\scriptsize Put the bowl on the plate} & \parbox[t]{0.24\textwidth}{\centering\scriptsize Put the bowl on top of the cabinet} & \parbox[t]{0.24\textwidth}{\centering\scriptsize Put the bowl on the stove} \\[6pt]
        \multicolumn{4}{c}{\textbf{Long}} \\[2pt]
        \includegraphics[width=0.24\textwidth]{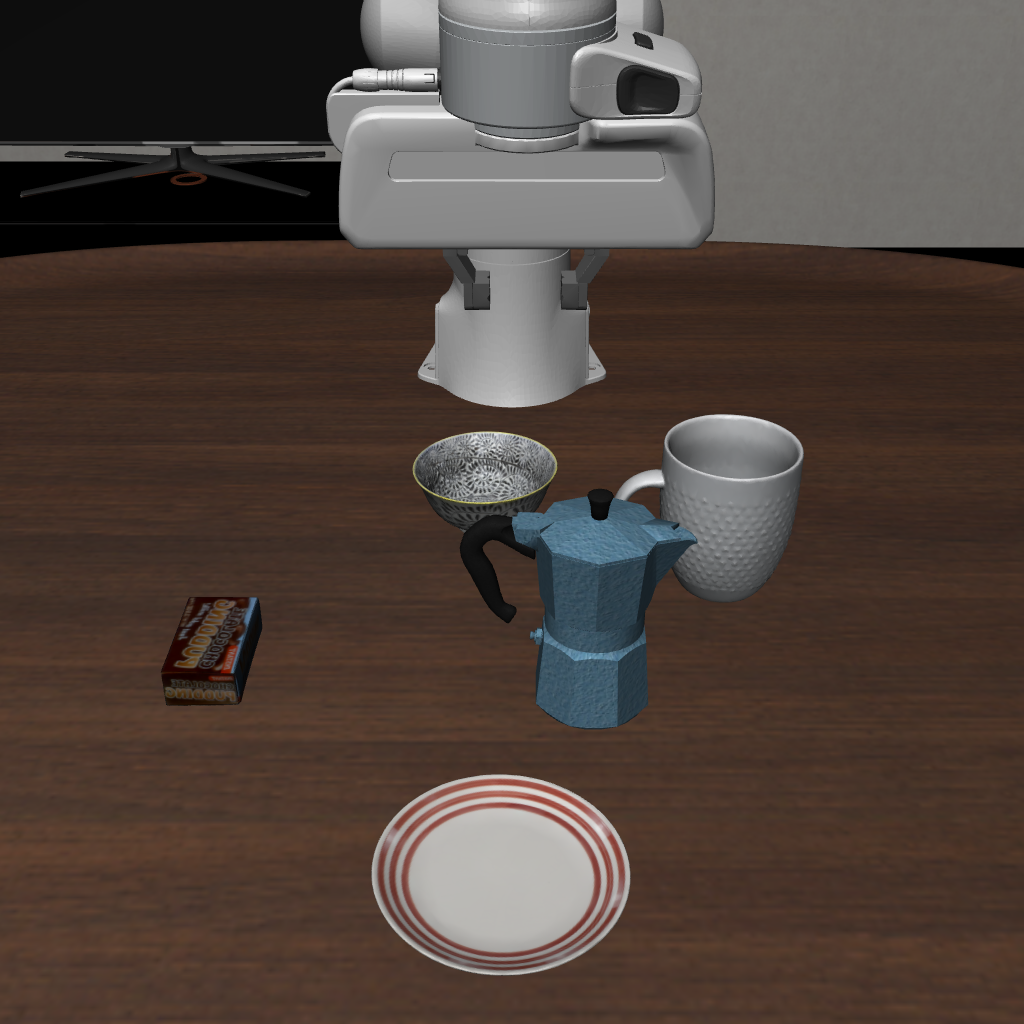} & \includegraphics[width=0.24\textwidth]{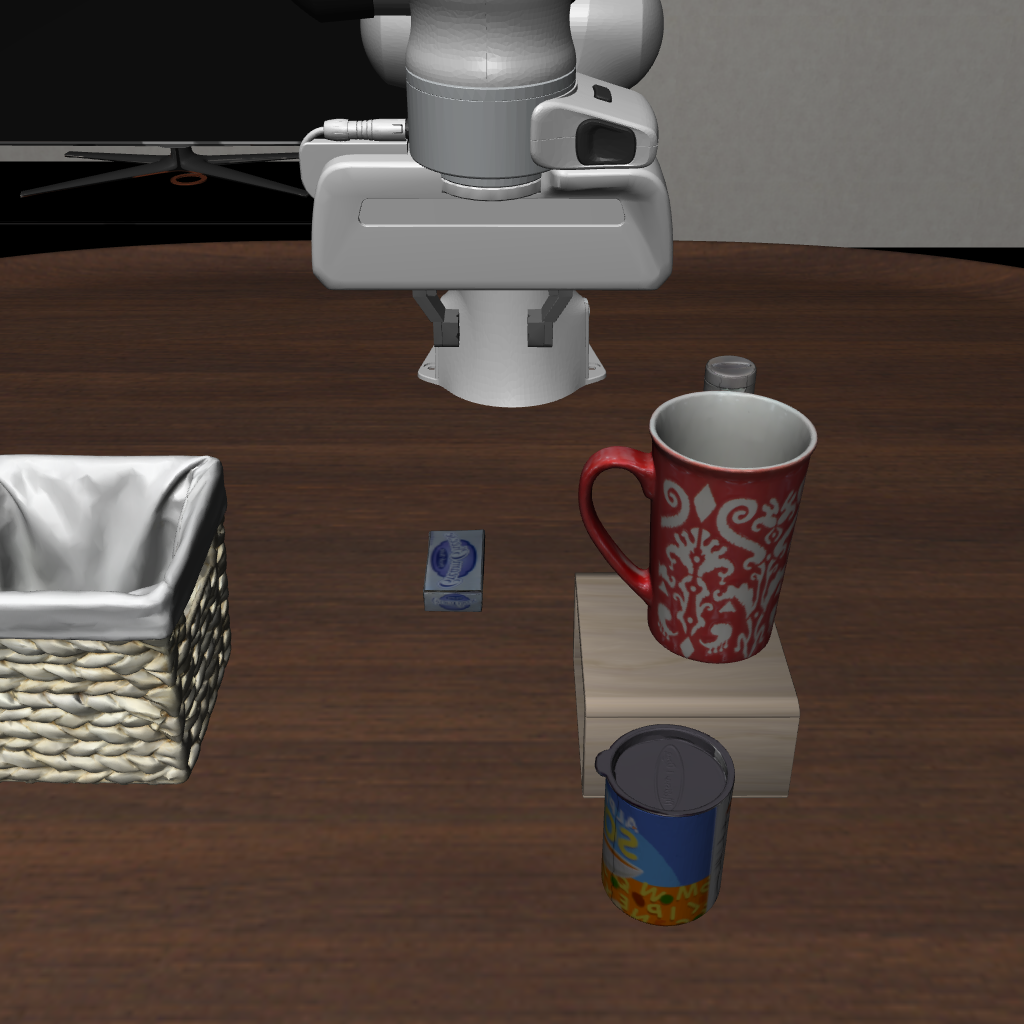} & \includegraphics[width=0.24\textwidth]{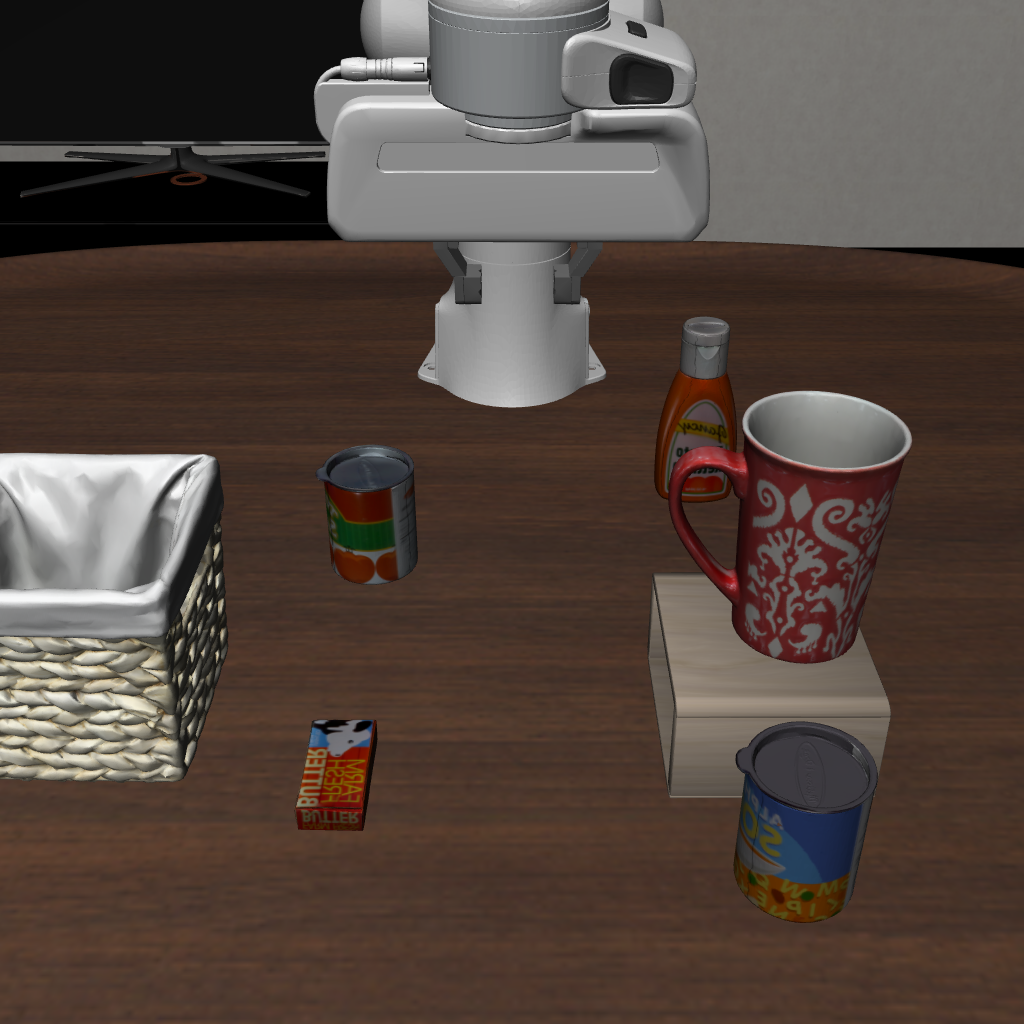} & \includegraphics[width=0.24\textwidth]{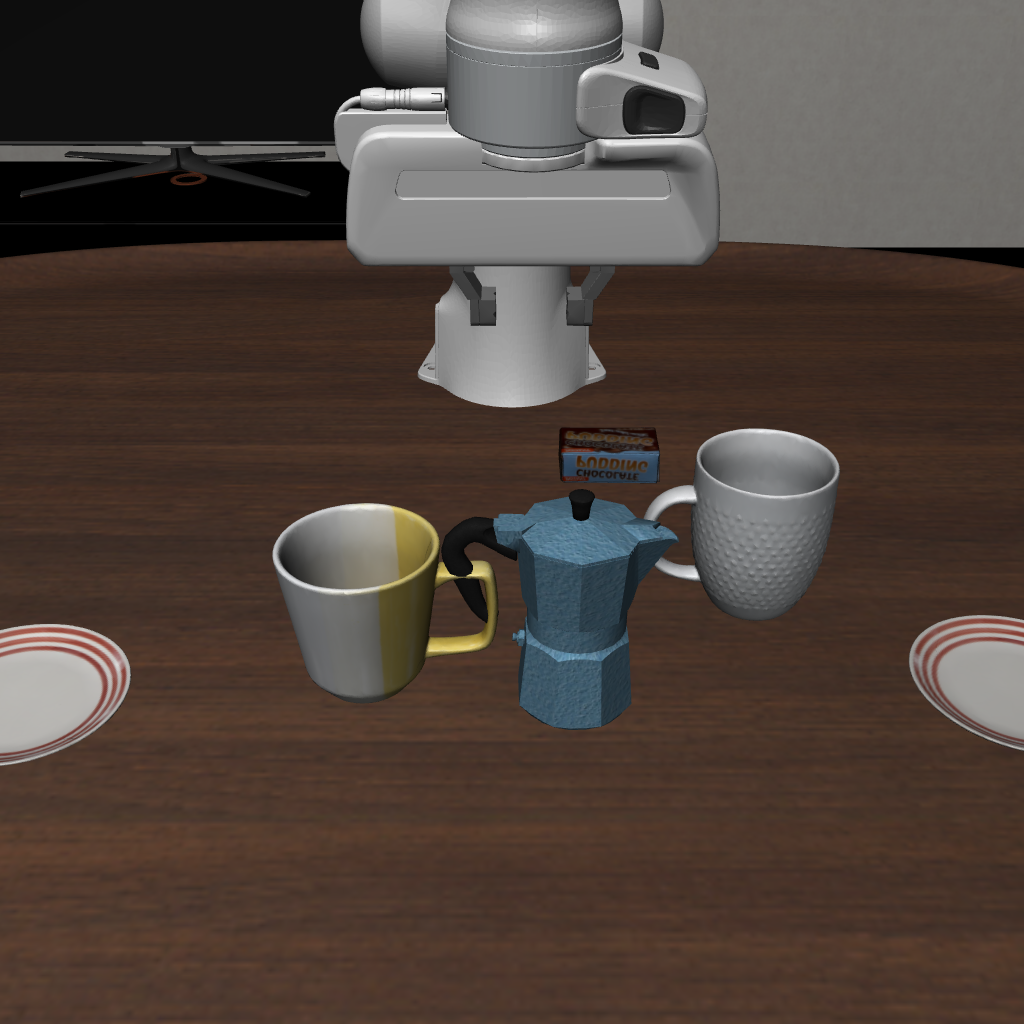} \\[1pt]
        \parbox[t]{0.24\textwidth}{\centering\scriptsize Put the white mug on the plate and put the chocolate pudding to the right of the plate} & \parbox[t]{0.24\textwidth}{\centering\scriptsize Put both the alphabet soup and the cream cheese box in the basket} & \parbox[t]{0.24\textwidth}{\centering\scriptsize Put both the alphabet soup and the tomato sauce in the basket} & \parbox[t]{0.24\textwidth}{\centering\scriptsize Put the white mug on the left plate and put the yellow and white mug on the right plate} \\
    \end{tabular}
    \caption{SafeLIBERO Safety Level II tasks. Four suites (Spatial, Object, Goal, Long) with obstacles at safety level ii density.}
    \label{fig:safelibero_tasks_ii}
\end{figure*}


\section{Additional Experimental Results}

\label{app:additional_results}
\begin{table*}[ht]
\centering
\caption{Expanded results across all SafeLIBERO tasks and safety levels.}
\label{tab:expanded-results-A}
\resizebox{\textwidth}{!}{
\label{tab:expanded_results}
\begin{tabular}{ll cc cc cc cc}
\toprule
& & \multicolumn{2}{c}{\textbf{Task 1}} & \multicolumn{2}{c}{\textbf{Task 2}} & \multicolumn{2}{c}{\textbf{Task 3}} & \multicolumn{2}{c}{\textbf{Task 4}} \\
\cmidrule(lr){3-4} \cmidrule(lr){5-6} \cmidrule(lr){7-8} \cmidrule(lr){9-10}
\textbf{Method} & \textbf{Metric} & S1 & S2 & S1 & S2 & S1 & S2 & S1 & S2 \\
\midrule

\multicolumn{10}{l}{\textit{\textbf{LIBERO-Spatial}}} \\
\midrule
\multirow{3}{*}{$\pi_{0.5}$}
& CAR ($\uparrow$)  & 2.00 & 0.00 & 10.00 & 52.00 & 20.00 & 2.00 & 36.00 & 0.00 \\
& TSR ($\uparrow$)  & 30.00 & 58.00 & 68.00 & 68.00 & 80.00 & 74.00 & 62.00 & 38.00 \\
& ETS ($\downarrow$)  & 253.60 & 192.70 & 190.60 & 189.30 & 178.50 & 187.60 & \textbf{176.40} & 244.50 \\
\cmidrule{1-10}
\multirow{3}{*}{AEGIS}
& CAR ($\uparrow$)  & \textbf{84.00} & 86.00 & 62.00 & \textbf{84.00} & 90.00 & 88.00 & \textbf{58.00} & 52.00 \\
& TSR ($\uparrow$)  & 50.00 & \textbf{88.00} & 74.00 & 80.00 & 90.00 & 84.00 & 48.00 & \textbf{72.00} \\
& ETS ($\downarrow$)  & 235.00 & \textbf{141.40} & 189.30 & 171.80 & \textbf{171.60} & 188.20 & 215.50 & 192.80 \\
\cmidrule{1-10}
\multirow{3}{*}{\textbf{Ours}}
& CAR ($\uparrow$)  & 80.00 & \textbf{90.00} & \textbf{80.00} & 66.00 & \textbf{96.00} & \textbf{90.00} & 54.00 & \textbf{56.00} \\
& TSR ($\uparrow$)  & \textbf{64.00} & 70.00 & \textbf{82.00} & \textbf{84.00} & \textbf{92.00} & \textbf{94.00} & \textbf{64.00} & 64.00 \\
& ETS ($\downarrow$)  & \textbf{197.90} & 188.10 & \textbf{173.90} & \textbf{167.60} & 214.40 & \textbf{177.30} & 184.30 & \textbf{187.30} \\

\midrule
\multicolumn{10}{l}{\textit{\textbf{LIBERO-Goal}}} \\
\midrule
\multirow{3}{*}{$\pi_{0.5}$}
& CAR ($\uparrow$)  & 0.00 & 46.00 & 12.00 & 62.00 & 0.00 & 32.00 & 18.00 & 20.00 \\
& TSR ($\uparrow$)  & 34.00 & 54.00 & 88.00 & 94.00 & 32.00 & 66.00 & 28.00 & 38.00 \\
& ETS ($\downarrow$)  & 257.10 & 220.10 & 141.70 & \textbf{99.30} & 267.70 & 187.80 & 277.60 & 231.00 \\
\cmidrule{1-10}
\multirow{3}{*}{AEGIS}
& CAR ($\uparrow$)  & \textbf{98.00} & 82.00 & \textbf{100.00} & \textbf{100.00} & 76.00 & 48.00 & \textbf{94.00} & 54.00 \\
& TSR ($\uparrow$)  & 80.00 & 64.00 & \textbf{92.00} & 96.00 & 54.00 & \textbf{78.00} & 72.00 & 66.00 \\
& ETS ($\downarrow$)  & 164.50 & 221.20 & \textbf{133.10} & 106.00 & 225.90 & \textbf{165.50} & 240.10 & \textbf{180.60} \\
\cmidrule{1-10}
\multirow{3}{*}{\textbf{Ours}}
& CAR ($\uparrow$)  & 96.00 & \textbf{88.00} & 92.00 & 86.00 & \textbf{98.00} & \textbf{72.00} & 90.00 & \textbf{84.00} \\
& TSR ($\uparrow$)  & \textbf{100.00} & \textbf{96.00} & \textbf{92.00} & \textbf{98.00} & \textbf{86.00} & 58.00 & \textbf{86.00} & \textbf{82.00} \\
& ETS ($\downarrow$)  & \textbf{105.10} & \textbf{156.50} & 143.50 & 154.40 & \textbf{219.60} & 279.70 & \textbf{239.00} & 362.60 \\

\bottomrule
\end{tabular}

\vspace{1mm}
\begin{minipage}{0.95\linewidth}
\end{minipage}
}
\end{table*}
\begin{table*}[ht]
\centering
\caption{Expanded results across all SafeLIBERO tasks and safety levels.}
\resizebox{\textwidth}{!}{
\label{tab:expanded-results-B}

\begin{tabular}{ll cc cc cc cc}
\toprule
& & \multicolumn{2}{c}{\textbf{Task 1}} & \multicolumn{2}{c}{\textbf{Task 2}} & \multicolumn{2}{c}{\textbf{Task 3}} & \multicolumn{2}{c}{\textbf{Task 4}} \\
\cmidrule(lr){3-4} \cmidrule(lr){5-6} \cmidrule(lr){7-8} \cmidrule(lr){9-10}
\textbf{Method} & \textbf{Metric} & S1 & S2 & S1 & S2 & S1 & S2 & S1 & S2 \\
\midrule
\multicolumn{10}{l}{\textit{\textbf{LIBERO-Object}}} \\
\midrule
\multirow{3}{*}{$\pi_{0.5}$}
& CAR ($\uparrow$)  & 42.00 & 12.00 & 0.00 & 56.00 & 0.00 & 52.00 & 8.00 & 14.00 \\
& TSR ($\uparrow$)  & 84.00 & 72.00 & 6.00 & 82.00 & 0.00 & \textbf{72.00} & 56.00 & 58.00 \\
& ETS ($\downarrow$)  & 175.90 & \textbf{184.20} & 297.60 & 193.40 & 300.00 & \textbf{173.30} & 248.50 & \textbf{211.20} \\
\cmidrule{1-10}
\multirow{3}{*}{AEGIS}
& CAR ($\uparrow$)  & 76.00 & \textbf{94.00} & 32.00 & \textbf{100.00} & \textbf{78.00} & \textbf{100.00} & 48.00 & 70.00 \\
& TSR ($\uparrow$)  & 88.00 & 74.00 & \textbf{92.00} & 92.00 & 74.00 & 68.00 & 82.00 & 72.00 \\
& ETS ($\downarrow$)  & 167.50 & 198.50 & \textbf{227.80} & \textbf{179.90} & \textbf{233.80} & 194.70 & \textbf{195.90} & 211.90 \\
\cmidrule{1-10}
\multirow{3}{*}{\textbf{Ours}}
& CAR ($\uparrow$)  & \textbf{84.00} & 64.00 & \textbf{100.00} & 92.00 & 46.00 & 92.00 & \textbf{100.00} & \textbf{94.00} \\
& TSR ($\uparrow$)  & \textbf{94.00} & \textbf{78.00} & 84.00 & \textbf{94.00} & \textbf{96.00} & 68.00 & \textbf{88.00} & \textbf{84.00} \\
& ETS ($\downarrow$)  & \textbf{154.40} & 314.60 & 399.20 & 301.70 & 298.30 & 484.30 & 225.90 & 266.60 \\

\midrule
\multicolumn{10}{l}{\textit{\textbf{LIBERO-Long}}} \\
\midrule
\multirow{3}{*}{$\pi_{0.5}$}
& CAR ($\uparrow$)  & 8.00 & 12.00 & 14.00 & 22.00 & 28.00 & 6.00 & 0.00 & 12.00 \\
& TSR ($\uparrow$)  & 42.00 & 36.00 & 24.00 & 36.00 & 52.00 & 22.00 & 50.00 & 24.00 \\
& ETS ($\downarrow$)  & 476.10 & 500.00 & 508.90 & \textbf{478.60} & 416.00 & 510.70 & \textbf{436.60} & 496.90 \\
\cmidrule{1-10}
\multirow{3}{*}{AEGIS}
& CAR ($\uparrow$)  & 78.00 & \textbf{94.00} & \textbf{92.00} & \textbf{98.00} & 84.00 & 56.00 & 58.00 & 77.00 \\
& TSR ($\uparrow$)  & 76.00 & 48.00 & 34.00 & 30.00 & 50.00 & 36.00 & 44.00 & 32.00 \\
& ETS ($\downarrow$)  & \textbf{424.40} & 486.00 & 514.90 & 513.50 & 440.30 & \textbf{489.10} & 477.10 & \textbf{495.70} \\
\cmidrule{1-10}
\multirow{3}{*}{\textbf{Ours}}
& CAR ($\uparrow$)  & \textbf{82.00} & 80.00 & 76.00 & 74.00 & \textbf{98.00} & \textbf{94.00} & \textbf{72.00} & \textbf{84.00} \\
& TSR ($\uparrow$)  & \textbf{84.00} & \textbf{90.00} & \textbf{74.00} & \textbf{70.00} & \textbf{94.00} & \textbf{62.00} & \textbf{74.00} & \textbf{66.00} \\
& ETS ($\downarrow$)  & 533.80 & \textbf{484.10} & \textbf{498.60} & 580.80 & \textbf{370.60} & 553.50 & 455.40 & 526.00 \\

\bottomrule
\end{tabular}
\vspace{1mm}
\begin{minipage}{0.95\linewidth}
\end{minipage}
}
\end{table*}

\end{document}